\pdfoutput=1

\documentclass[11pt]{article}
\usepackage[table]{xcolor} 

\usepackage[]{acl}

\usepackage{times}
\usepackage{latexsym}
\usepackage{todonotes}
\usepackage[T1]{fontenc}

\usepackage[utf8]{inputenc}
\usepackage{enumitem}

\usepackage{pgfplots}
\usepackage{xcolor}
\usepackage{tikz}
\usepackage{bchart}

\usepackage{placeins} 

\definecolor{afrispeechcolor}{HTML}{70FFB3}
\definecolor{alffacolor}{HTML}{70D6FF}
\definecolor{bembacolor}{HTML}{FF70A6}
\definecolor{biblecolor}{HTML}{FF9770}
\definecolor{cv19color}{HTML}{FFD670}
\definecolor{finspeechcolor}{HTML}{E9FF70}
\definecolor{kallaamacolor}{HTML}{a5ffd6}
\definecolor{kinyacolor}{HTML}{ff69eb}
\definecolor{lwazicolor}{HTML}{7371fc}
\definecolor{naijacolor}{HTML}{D62728}
\definecolor{nchtlcolor}{HTML}{ff686b}
\definecolor{nicolingua3color}{HTML}{9467BD}
\definecolor{nicolingua4color}{HTML}{C5B0D5}
\definecolor{ologoafricacolor}{HTML}{ffe45e}
\definecolor{sattcolor}{HTML}{8C564B}
\definecolor{soapcolor}{HTML}{C49C94}
\definecolor{spcscolor}{HTML}{E377C2}
\definecolor{tunswitchcscolor}{HTML}{F7B6D2}
\definecolor{tunswitchtocolor}{HTML}{7F7F7F}
\definecolor{tunswitchweakaudiocolor}{HTML}{C7C7C7}
\definecolor{udhrcolor}{HTML}{C7C7C7}
\definecolor{voacolor}{HTML}{BCBD22}
\definecolor{voxcolor}{HTML}{DBDB8D}
\definecolor{yourbavoicecolor}{HTML}{17BECF}
\definecolor{zambezi1color}{HTML}{9EDAE5}
\definecolor{zambezi2color}{HTML}{fe5d9f}

\newcommand{\dsbox}[1]{\textcolor{#1}{\rule{1.5ex}{1.5ex}}}
\newcommand{\dsentry}[2]{(\dsbox{#1}, #2)}
\newcommand{\dsentrycs}[3]{(\dsbox{#1}, #2, #3)}

\newcommand{\sttag}{\begingroup\fboxsep=1.5pt\fboxrule=0.5pt\fcolorbox{black!20}{blue!10}{\tiny ST}\endgroup}
\newcommand{\tstag}{\begingroup\fboxsep=1.5pt\fboxrule=0.5pt\fcolorbox{black!20}{orange!10}{\tiny TS}\endgroup}
\newcommand{\ldtag}{\begingroup\fboxsep=1.5pt\fboxrule=0.5pt\fcolorbox{black!20}{green!10}{\tiny LD}\endgroup}
\newcommand{\pttag}{\begingroup\fboxsep=1.5pt\fboxrule=0.5pt\fcolorbox{black!20}{purple!10}{\tiny PT}\endgroup}

\usetikzlibrary{shapes.geometric}
\definecolor{ptcolor}{HTML}{FFD700}  
\definecolor{ftcolor}{HTML}{17BECF}  
\newcommand{\dsdiamond}[1]{%
  \tikz[baseline=-1ex, x=0.75ex, y=0.75ex]
    \draw[fill=#1, draw=#1] 
      (0,0.7) -- (0.7,0) -- (0,-0.7) -- (-0.7,0) -- cycle;%
}

\newcommand{\dsft}[1]{\dsdiamond{ftcolor}} 

\usepackage{booktabs}
\usepackage{threeparttable}

\definecolor{findOptimalPartition}{HTML}{D7191C}
\definecolor{storeClusterComponent}{HTML}{FDAE61}
\definecolor{dbscan}{HTML}{ABDDA4}
\definecolor{constructCluster}{HTML}{2B83BA}

\usepackage{subfig} 
\usepackage{microtype}
\usepackage{graphicx}
\usepackage{booktabs} 
\usepackage{tabularx}
\usepackage{multirow} 
\usepackage{hyperref}
\usepackage{lipsum}
\usepackage{xltabular}
\usepackage{longtable}
\usepackage{cals}
\usepackage{xcolor,colortbl}
\usepackage[normalem]{ulem}
\usepackage{pifont}

\def \ourbenchmark{Sahara}

\usepackage{soul}

 \usepackage{multirow}
\usepackage{amsmath}
\usepackage{amssymb}
\usepackage{listings}
\usepackage{algorithm}
\usepackage{algpseudocode}

\usepackage{appendix}
\usepackage{array} 
\newcolumntype{H}{>{\setbox0=\hbox\bgroup}c<{\egroup}@{}}

\usepackage[normalem]{ulem}  
\usepackage{soul}            

\newcommand{\linecolor}[2]{\setulcolor{#1}\ul{#2}}

\usepackage{hyperref} 

\usepackage{xspace}
\def \ourmodel{\textit{Simba}\xspace}
\def \ourbenchmark{\textit{SimbaBench}\xspace}
\def \numSpeechHours{8,605}

\usepackage{pgfplots}
\pgfplotsset{compat=1.18}

\title{\includegraphics[scale=0.02]{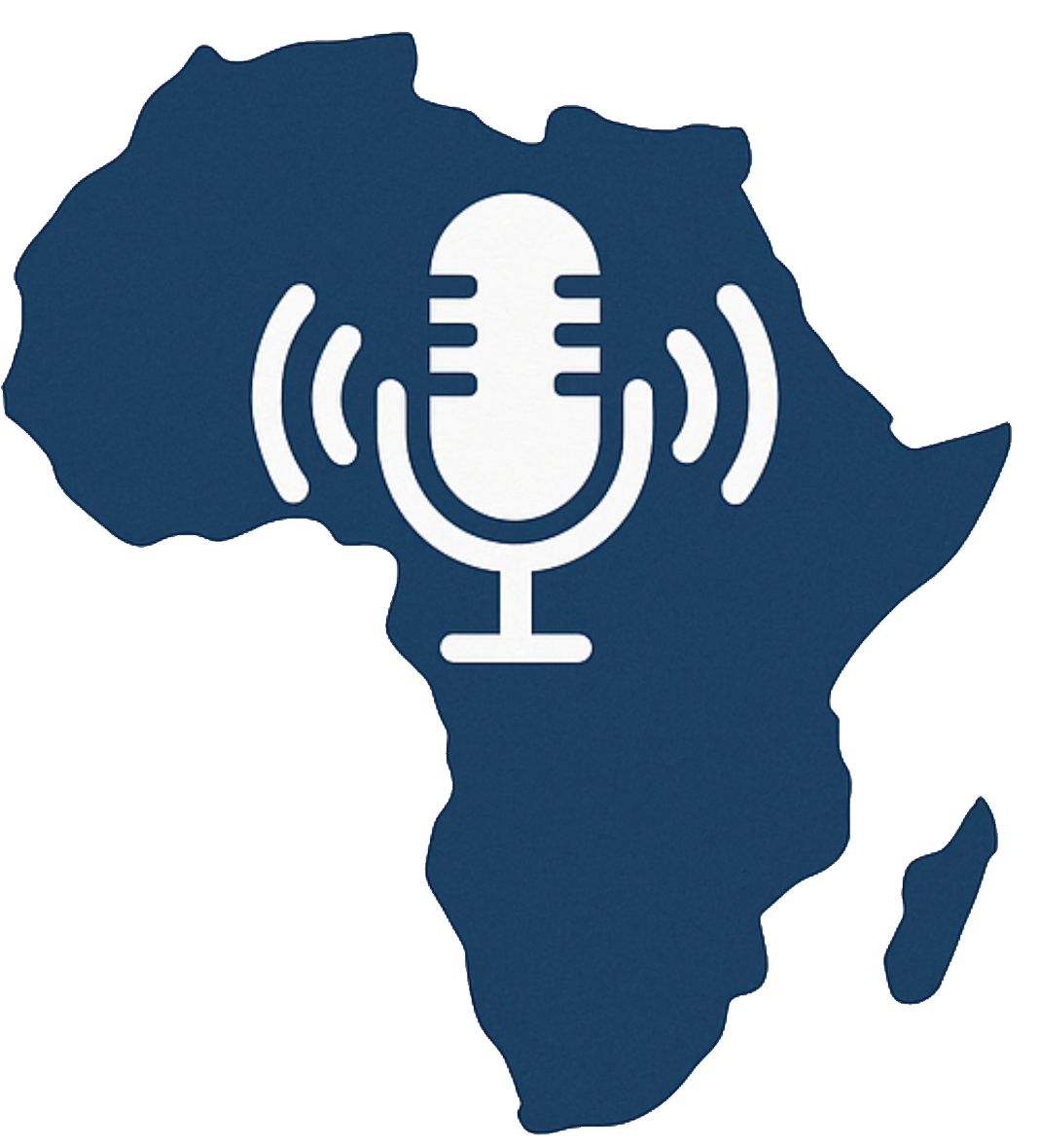} Voice of a Continent: Mapping Africa’s Speech Technology Frontier}

\author{\begin{tabular}[c]{@{}c@{}}
\normalsize  AbdelRahim Elmadany$^{\xi}$ ~  Sang Yun Kwon$^{\xi}$ ~  Hawau Olamide Toyin$^{\Omega}$ \\
\normalsize  ~Alcides Alcoba Inciarte$^{\xi}$ ~ Hanan Aldarmaki$^{\Omega}$ ~ Muhammad Abdul-Mageed$^{\xi,\lambda}$\end{tabular}\\
\normalsize $^{\xi}$The University of British Columbia ~~~~~ $^{\Omega}$MBZUAI ~~~~~ $^\lambda$ Invertible AI\\ %
  \texttt{\normalsize \{a.elmadany,muhammad.mageed\}@ubc.ca}}

\begin{document}
\maketitle

\begin{abstract}


Africa's rich linguistic diversity remains significantly underrepresented in speech technologies, creating barriers to digital inclusion. To alleviate this challenge, we systematically map the continent's speech space of datasets and technologies, leading to a new comprehensive benchmark \ourbenchmark for downstream African speech tasks. Using \ourbenchmark, we introduce the \ourmodel family of models, achieving state-of-the-art performance across multiple African languages and speech tasks. Our benchmark analysis reveals critical patterns in resource availability, while our model evaluation demonstrates how dataset quality, domain diversity, and language family relationships influence performance across languages. Our work highlights the need for expanded speech technology resources that better reflect Africa's linguistic diversity and provides a solid foundation for future research and development efforts toward more inclusive speech technologies.

\end{abstract}

\section{Introduction}\label{sec:introduction}

\begin{figure}[!ht]
  \centering
  \includegraphics[width=0.98\columnwidth]{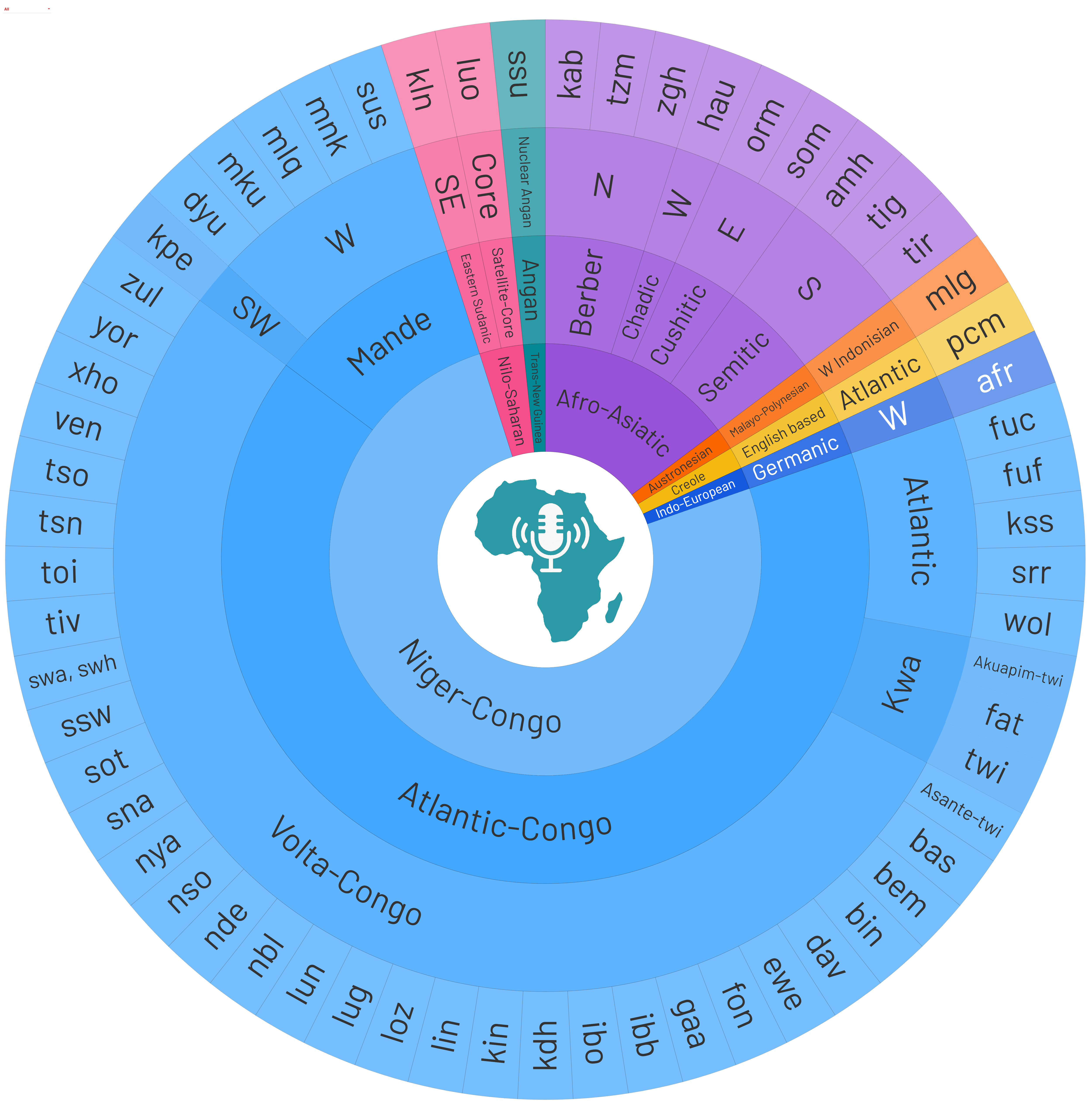}
\caption{A three-level language family hierarchy illustrating the 61 African languages included in our analysis, benchmark, and speech modeling efforts.}
\label{fig:sunburst} 
\end{figure}


Speech is one of the most natural and fundamental forms of human communication. Advances in speech technologies, such as automatic speech recognition (ASR), text-to-speech (TTS), and spoken language understanding, have enabled transformative applications including virtual assistants, real-time translation, and accessible communication tools for people with disabilities. However, the benefits of these technologies are not equitably distributed. Most current resources and research efforts are concentrated on a handful of widely spoken languages, particularly English, leaving the majority of the world’s linguistic diversity underrepresented~\cite{bender2011language, joshi2020state}. This imbalance is especially stark in the context of African languages, which are spoken by hundreds of millions but often lack the data and tools necessary for the development of robust speech systems. Addressing this gap is crucial for fostering technological inclusion, preserving linguistic heritage, and enabling culturally relevant digital innovation. Moreover, as large language models (LLMs) increasingly integrate speech capabilities~\cite{huang2024audiogpt, cui2024recent, nguyen2023generative, nguyen2025spirit}, ensuring that African languages are supported in both text and speech modalities is essential for equitable access to emerging AI technologies.

While recent multilingual speech models such as Whisper~\cite{whisper}, MMS~\cite{pratap2023mms}, and SeamlessM4T~\cite{seamlessm4t2023} include some coverage of African languages, their performance on key speech tasks such as ASR, TTS, and spoken language identification (SLID) remains inadequate, especially for low-resource and tonal languages. Despite recent efforts to improve speech modeling for African languages like mHuBERT~\cite{mhubert} and AfriHUBERT~\cite{alabi2024afrihubert}, these models cover only a small fraction of Africa's languages. 
In addition, African speech datasets are often undocumented or fragmented, with little clarity on their scope, supported tasks, language coverage, and evaluation standards.

\begin{figure}[ht]
  \centering
    \includegraphics[width=0.8\linewidth]{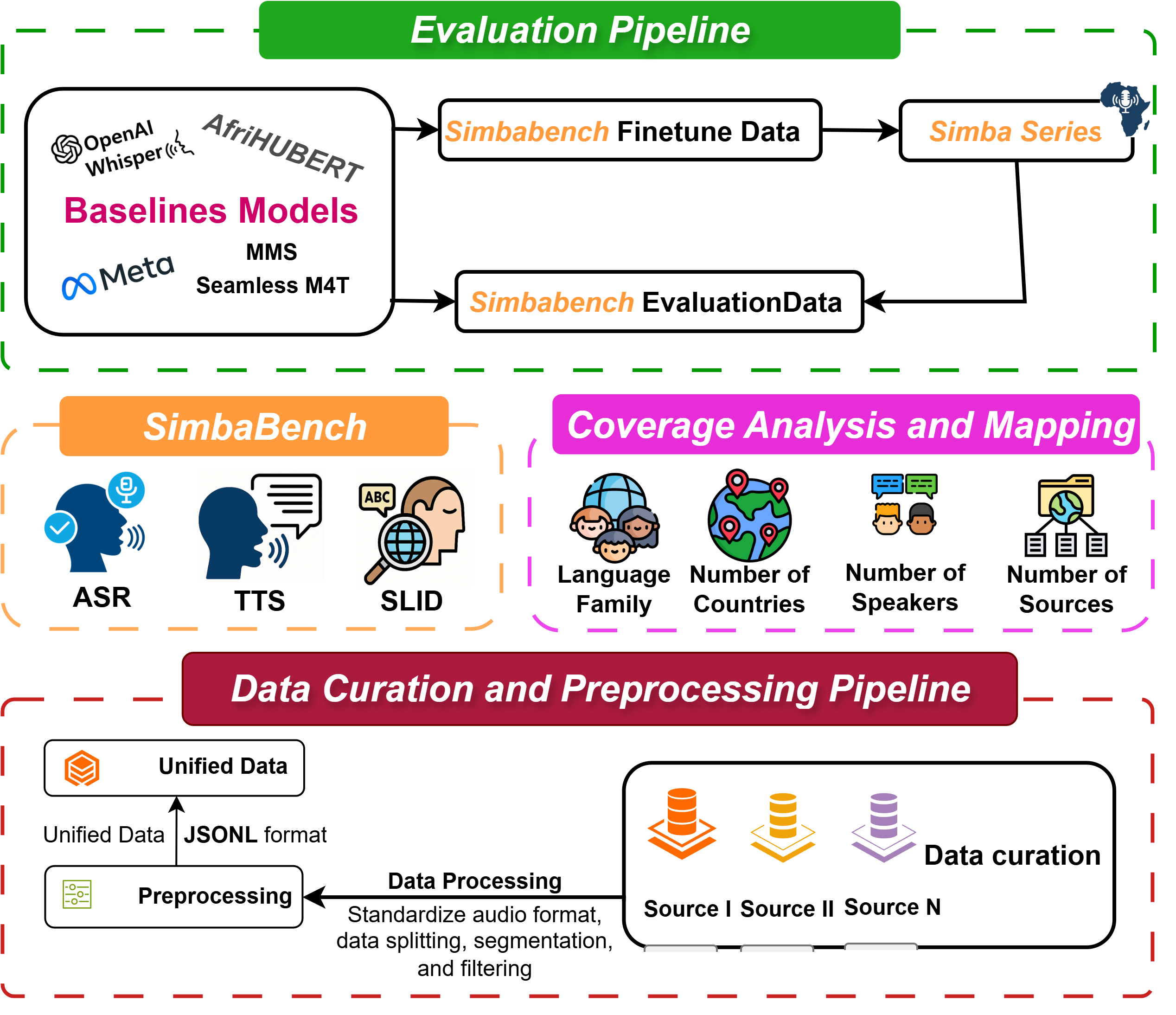}
  \caption{Methodological workflow, illustrating the three main components: (1) the data curation and preprocessing pipeline, (2) \ourbenchmark with quantitative mapping of current speech datasets in Africa, and (3) the evaluation pipeline.}
  \label{fig:data_workflow}
\end{figure}

Recognizing the critical need to clearly characterize the current landscape of African speech datasets and technologies, we undertake a mapping of these resources and systems. In particular, we offer a number of contributions: \textbf{(1) New Speech Benchmark:} we conduct extensive data collection and aggregate and harmonize all publicly available resources covering ASR, TTS, and SLID tasks. This dataset collection spans diverse linguistic families and geographic regions, leading the way to the development of \textbf{\ourbenchmark}, a unified benchmark designed specifically for African speech processing. \textbf{(2) Data-Driven Coverage Analysis:} with \ourbenchmark at hand, we carry out a quantitative mapping of current speech datasets in Africa, allowing us to draw connections between dataset availability across languages and populations. This helps paint the picture for the current state of African speech resources. \textbf{(3) Model Evaluation:} we benchmark existing state-of-the-art (SoTA) African and multilingual speech models on \ourbenchmark, thereby empirically assessing capabilities and limitations of these models across African speech tasks. These evaluations offer critical insights into where current models fall short and where targeted innovation is needed. \textbf{(4) A Family of SoTA African Speech Models:} we exploit our datasets to build upon existing models, introducing a suite of fine-tuned models, dubbed \textit{Simba}, achieving SoTA performance on a wide set of African languages across the downstream tasks. Figure~\ref{fig:data_workflow} illustrates the methodological workflow employed in our work.

Through this work, we provide foundational tools and resources to accelerate speech technology for African languages and invite community participation in this inclusive, multilingual effort. The paper is organized as follows: Section~\ref{sec:litreview} overviews related work in African NLP and Speech. Section~\ref{sec:data} describes our data mapping and collection process. We detail our benchmark \ourbenchmark in Section~\ref{subsec:data_simbabench}. Section~\ref{sec:eval_othersmodel_evaluation} outlines our evaluation setup, and we discuss results and findings in Sections~\ref{sec:result} and~\ref{sec:discussion}.

\section{Literature Review}\label{sec:litreview} 
\noindent Speech and language technologies enable broader access to information and can potentially support and promote linguistic diversity. However, of the over 7,000 languages spoken worldwide, only a select few are represented in contemporary language technologies and applications~\cite{joshi2020state}. Most speech and NLP systems are predominantly trained on a limited subset of languages, primarily from dominant language families and specific geographies, leaving most languages unrepresented~\cite{ponti-etal-2019-modeling}. \citet{joshi2020state} categorizes languages into 6 classes based on available resources, ranging from \textbf{\textit{The Left-Behinds}} (Class 0) with virtually no digital presence to \textbf{\textit{The Winners}} (Class 5) with abundant resources and technological support. With most African languages occupying the lower tiers of this classification, a substantial language gap persists, leaving indigenous and regional languages under-represented in NLP~\cite{adebara2025sahara}, highlighting the need for additional efforts to promote inclusive language technologies~\cite{ojo2023good}. 

\paragraph{Progress in African NLP.} In recent years, significant progress has been made towards improving representation and performance of African languages in NLP, particularly in text understanding and generation tasks~\cite{adebara2022towards,adebara2025sahara}. Benchmarks like SAHARA~\cite{adebara2025sahara}, IrokoBench~\cite{adelani2024irokobench}, and others~\cite{ojo2023good, wang2023afrimte, oladipo-etal-2023-better, reid2021afromt} have advanced NLU and NLG capabilities. In terms of model development, models like AfroXLMR~\cite{belay2025afroxlmr}, Cheetah~\cite{adebara2024cheetah} and others have contributed significantly to these developments~\cite{adebara2022serengeti, elmadany2024toucan, adebara-etal-2022-afrolid}. Despite these advancements, the development of African speech technologies remains slow, impeded by intenstive computational requirements, a shortage of large-scale speech corpora, and historical bias towards high-resource Western languages~\cite{joshi2020state}. This resource gap motivates our work toward inclusive technologies for Africa's diverse languages.

\paragraph{Progress in African Speech.} Prior works on African speech involve mostly speech resource collection and presentation of baseline results~\cite{ogun20241000,gutkin-et-al-yoruba2020,sikasote23_interspeech, meyer2022bibletts}. Although large scale speech models are becoming increasingly multilingual, majority of African languages are left behind, with most African languages excluded during pretraining for large speech models~\cite{alabi2024afrihubert}. Despite these limitations, models like AfriHuBERT~\cite{alabi2024afrihubert} exemplify recent efforts to address this gap.

\section{Mapping the Data Landscape}\label{sec:data}

\begin{table}[!ht]
\centering
\renewcommand{\arraystretch}{1.1}
\resizebox{\columnwidth}{!}{%
\begin{tabular}{cHlcrl}
\toprule
\textbf{Task} & \textbf{Type} & \textbf{Dataset} & \textbf{\#Lang.} & \textbf{Dur. (h)} & \textbf{Domain} \\
\midrule

\multirow{14}{*}{\rotatebox{90}{\textbf{ASR}} } &\multirow{11}{*}{Normal}  & Alffa Public~\cite{ALFFA_PUBLIC} & 4 & 58.66 & RS, N \\
   &   & BembaSpeech~\cite{sikasote-anastasopoulos:2022:LREC} & 1 & 26.93 & N, V \\
   &    &Common Voice (CV-19)~\cite{commonvoice} & 21 & 1,843.65 & RS \\
   &    &Financial Speech~\cite{FinancialInclusionSpeechDataset} & 4 & 149.55 & RS, F \\
   &    &Kallaama~\cite{Gauthier2024Kallaama} & 3 & 113.68 & R, IR \\
   &    &Lwazi~\cite{vanHeerden2016LwaziASR} & 10 & 42.80 & TC \\
   &    &Naija Voices~\cite{NaijaVoices2024} & 3 & 1,867.52 & RS \\
   &    &NCHLT + AUX1/2~\cite{Barnard2014NCHLT} & 11 & 1,922.05 & RS \\
   &    &Nicolingua (0004)~\cite{doumbouya2021usingradio} & 3 & 1.24 & R \\
   &    &YorubaVoice~\cite{gutkin-et-al-yoruba2020} & 1 & 4.03 & G \\
   &    &Zambezi Voice (ASR)~\cite{sikasote23_interspeech} & 3 & 54.23 & RS, TS \\ \cmidrule{2-6}
      & \multirow{2}{*}{Code-Switched}  & SO (Code-Switched)~\cite{VAN_DER_WESTHUIZEN18.90} & 4 & 14.27 & TV \\
    &    &SPCS (Code-Switched)~\cite{Modipa2015CodeSwitching} & 1 & 10.48 & R\\ \cmidrule{2-6}

         
                  & & \textbf{ASR Statistics} & \textbf{42} & \textbf{6,109.09} &  \\

\midrule
\multirow{6}{*}{\rotatebox{90}{\textbf{SLID}}}  &--- & Nicolingua (0003)~\cite{doumbouya2021usingradio}  & 6 & 143.75 & R \\
   & New (\textcolor{red}{Ours})  & OlongoAfrica (\textcolor{red}{Ours})& 10 & 2.40 & SS \\   
   & New (\textcolor{red}{Ours})  &UDHR (\textcolor{red}{Ours})& 6 & 1.05 & HR \\
   &New (\textcolor{red}{Ours}) & Voice of Africa (VOA) (\textcolor{red}{Ours})& 10 & 865.08 & N \\
   &--- & VoxLingua~\cite{Valk2021VoxLingua107} & 9 & 773.66 & V \\
   & ---&Zambezi Voice (Audio Only)~\cite{sikasote23_interspeech} & 5 & 176.00 & TS \\ \cmidrule{2-6}
         
                  & & \textbf{SLID Statistics} & \textbf{39} & \textbf{1,961.94} &  \\

\midrule
\multirow{3}{*}{\rotatebox{90}{\textbf{TTS}}}  & & BibleTTS~\cite{meyer2022bibletts} & 6 & 306.69 & RB \\
   & ---&High-Quality TTS (SA)~\cite{van-niekerk-etal-2017} & 4 & 13.16 & WS \\
   & --- &Kinyarwanda TTS ~\cite{DigitalUmuganda_afrispeak_2023} & 1 & 14.08 & --- \\\cmidrule{2-6}
         
                  & & \textbf{TTS Statistics} & \textbf{11} & \textbf{333.93} &  \\ \midrule

       
 & --- & AfriSpeech (Accented-African))~\cite{olatunji2023afrispeech} & 1 & 200 & C, G \\
\midrule
   & &\textbf{Overall} & \textbf{61} &\textbf{8,604.96} & --- \\ \bottomrule


\end{tabular}
}

\caption{Overview of curated African audio datasets used in our data. This summary includes dataset type, number of languages covered (\textit{\#Lang.}), total duration in hours (\textit{Dur.}), and source domain. ``\textcolor{red}{Ours}'' refer to new data that we primarily collected or curated as part of this work. \textbf{RS.} refers to Read Speech, \textbf{TS.} Talk Show \textbf{TC.} Telephone Conversations, \textbf{F.} Financial, \textbf{TV.} TV Shows, \textbf{IR.} Interviews, \textbf{N.} News, \textbf{C.} Clinical, \textbf{SS.} Short Stories, \textbf{G.} General, \textbf{HR.} Human Rights, \textbf{R.} Radio, \textbf{V.} Video, \textbf{SO.} Soap Opera, \textbf{WS.} Wikipedia-based Speech, and \textbf{RB} Read Bible.} 

\label{tab:speech_data}
\end{table}

To understand the current state of speech technology for African languages, we begin with a comprehensive assessment of the available data resources. This is a necessary step before evaluating the capabilities of current models or proposing new directions for building robust multilingual systems. In particular, it is essential to identify what resources are available, where they originate, and where critical gaps persist. Below, we present an overview of publicly available African speech corpora, encompassing both labeled and unlabeled audio data. Our analysis centers on three core speech downstream tasks: ASR, TTS, and SLID.

Our objective is beyond mere data collection. Rather, our aim is to map the linguistic and acoustic diversity represented within existing datasets. This mapping lays the groundwork for a comprehensive and inclusive data infrastructure that authentically represents the multilingual realities of the African continent, which, as emphasized by~\citet{adebara2025sahara}, is crucial for ensuring equitable participation in global language technology advancements. 


\subsection{Data Curation}\label{subsec:speech_curation}


We curate a large-scale corpus of publicly available audio data integrating both labeled and unlabeled speech to ensure broad linguistic, acoustic, and demographic coverage. In total, we aggregate $\numSpeechHours$ hours of audio drawn from $26$ publicly available sources, comprising well-established corpora for downstream tasks (supervised) as well as large-scale unlabeled speech data (unsupervised). The collected resources span multiple domains, including media-rich and culturally grounded sources. This introduces variability in speech styles, regional dialects, and speaker identities—dimensions often underrepresented in traditional benchmarks. 


\noindent\textbf{African Data.} We collect over $8,380$ hours of clean data spanning $61$ african languages. Consisting of richly diverse domains like, \textit{broadcast}, \textit{radio}, \textit{read speech}, and \textit{spontaneous conversations}. This includes $6,080$ hours of ASR covering $42$ languages, $334$ hours of TTS spanning $11$ languages, and $1,960$ hours of untranscribed (audio-only) data across $32$ languages for SLID.

\noindent\textbf{Code-switched Data.} We include $\sim34$ hours of code-switched speech data, encompassing seven language pairs that combine African and non-African languages within a single utterance. These recordings reflect authentic patterns of multilingual discourse in everyday African contexts and are essential for training models capable of handling spontaneous, mixed-language input.


\noindent\textbf{African-accented English.} Furthermore, we incorporate $200$ hours of African-accented English speech, representing $120$ distinct accents from $13$ African countries with $2,463$ unique speakers~\cite{olatunji2023afrispeech}. 


A comprehensive summary of the dataset composition, language distribution, and task coverage is presented in Table~\ref{tab:speech_data}, with additional details provided in Appendix~\S\ref{app_sec:data}. Also, Table~\ref{appdex_tab:iso_codes_table} (in Appendix \S\ref{appdx_sec:mapping}) provides detailed information on the total audio duration (in hours) for each language across various datasets.

\begin{figure*}[ht]
  \centering
  \subfloat[Audio hours per language.]{
    \includegraphics[width=0.37\textwidth]{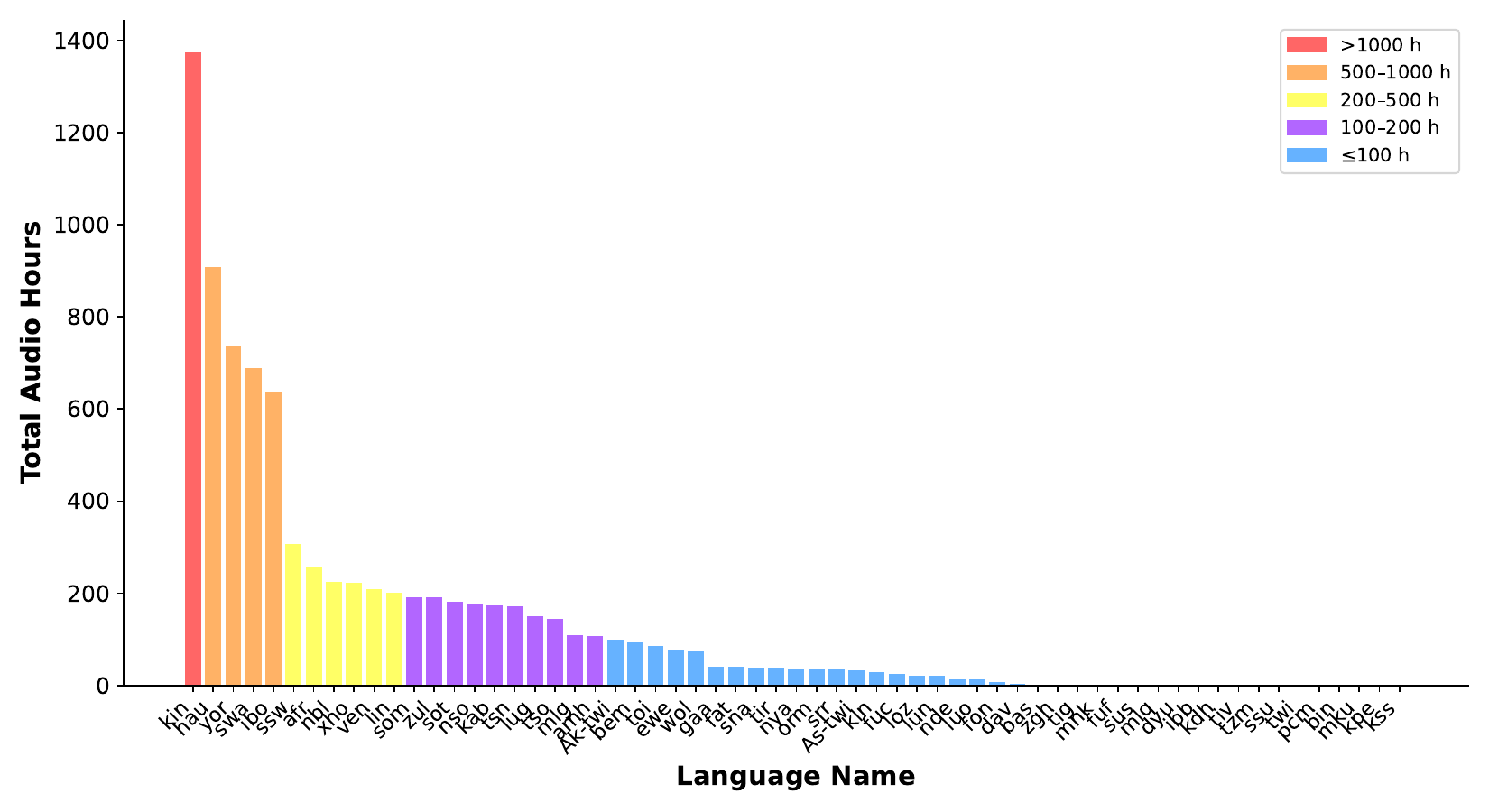}
    \label{fig:distrib_analysis_a}
  }
  \hfill
  \subfloat[Distribution density (KDE).]{
    \includegraphics[width=0.27\textwidth]{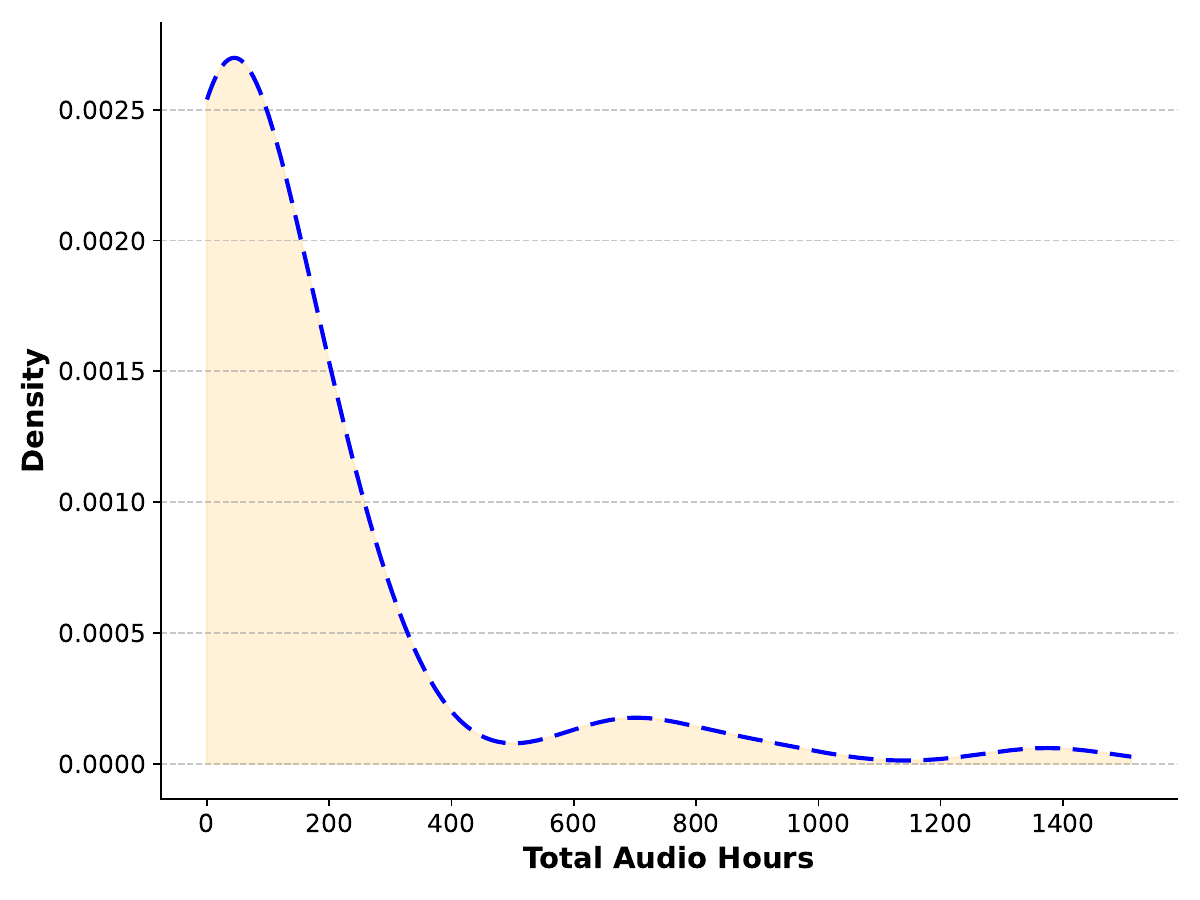}
    \label{fig:distrib_analysis_b}
  }
  \hfill
  \subfloat[Hours vs. number of sources.]{
    \includegraphics[width=0.25\textwidth]{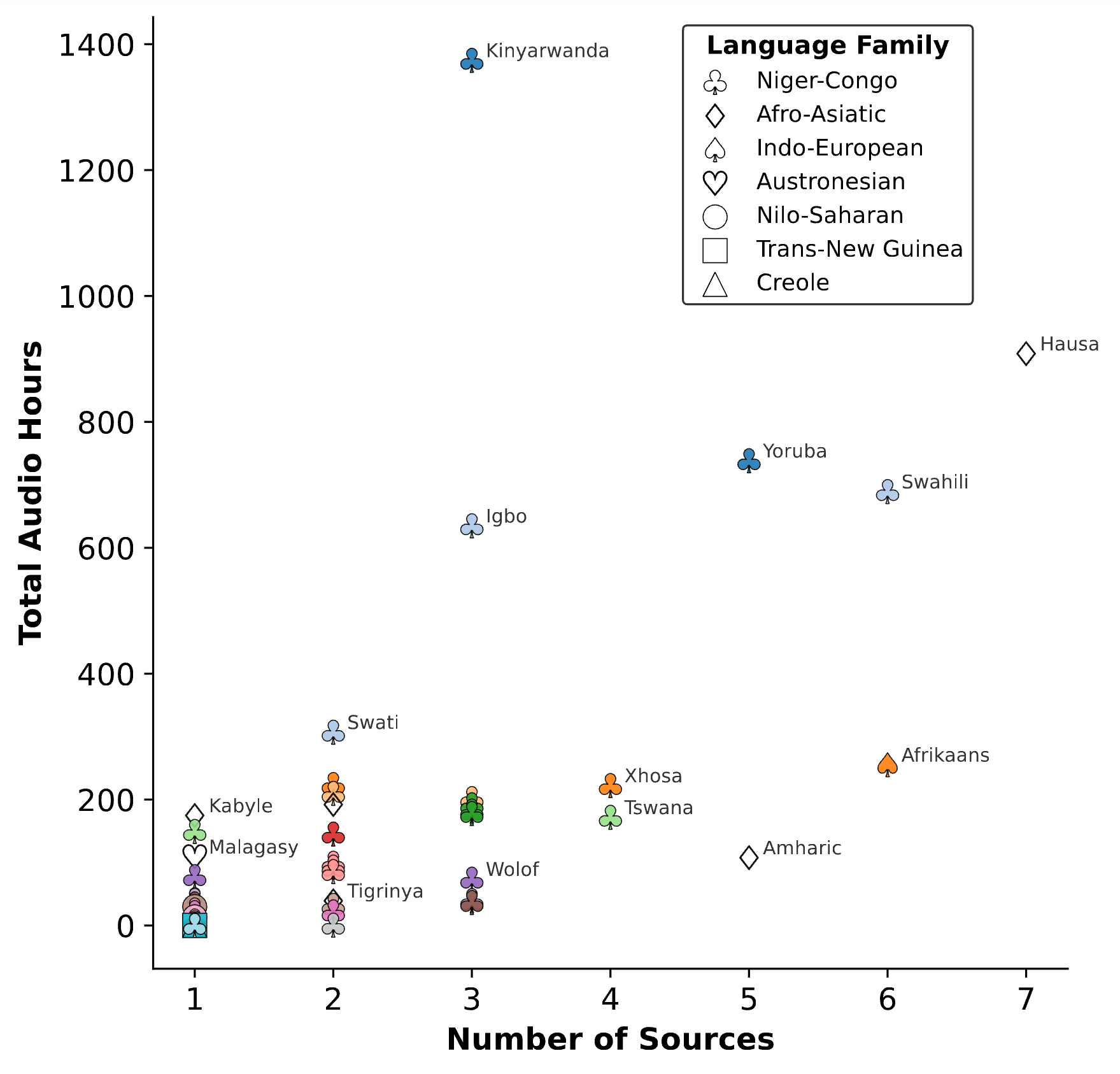}
    \label{fig:distrib_analysis_c}
  }
  \caption{Speech data distribution across the 61 African languages in collected data, highlighting volume, density, and source diversity.}
  \label{fig:distrib_analysis}
\end{figure*}




\subsection{Data Preprocessing and Standardization}\label{subsec:data_preprocessing}
To ensure consistency, quality, and usability across the diverse audio datasets, we apply a unified preprocessing pipeline encompassing \textit{format standardization}; converting all audio to 16 kHz mono WAV format, \textit{segmentation}, \textit{filtering}, and \textit{noise removal}; breaking long recordings into 1-20 second utterances and eliminating excessive noise, and \textit{metadata consolidation}; reformatting datasets into a unified JSON schema with standardized fields. Our preprocessing pipeline enables robust training and evaluation across diverse African speech corpora, establishing a foundation for consistent benchmarking and inclusive model development across all downstream tasks. More detailed information is outlined in Appendix~\S\ref{app_sec:preprocessing_pipeline}.

\subsection{Quantitative  Data Analysis}\label{subsec:data_analysis}
We present a quantitative analysis of the curated audio resources with respect to language distribution, task-specific coverage, and overall data volume. Our findings reveal disparities in speech data availability across African languages. We highlight the strengths and limitations of current African speech datasets, informing the feasibility of training and evaluating models for the aforementioned tasks. Together, these trends underscore the need for strategic data collection that prioritizes not only volume but also domain diversity and equitable representation across linguistic and demographic factors. Without such targeted efforts, existing disparities for African speech technology development will likely persist or worsen, further marginalizing already under-resourced languages. 

\paragraph{Overall Data Distribution.} Africa has more than $2000$ languages and dialects, of which our extensive efforts could only identify $61$ that have publicly available data. Even within this small number of languages, we find a minority of languages—\textit{Kinyarwanda}, \textit{Hausa}, \textit{Yoruba}, \textit{Swahili}, and \textit{Igbo}—accounting for hundreds to thousands of hours of recorded speech, whereas the majority of languages have fewer than one hour of data as shown in Figure~\ref{fig:distrib_analysis_a}. Figure~\ref{fig:distrib_analysis_b} highlights this imbalance through the Kernel Density Estimate (KDN) of total hours collected, revealing a heavily right‑skewed distribution with high density concentrated near zero hours and a long tail extending toward the few resource‑rich languages. This pattern highlights the imbalance driven primarily by targeted collection efforts rather than linguistic or demographic representation.

\paragraph{Language Family Distribution.} The distribution by language family in Figure~\ref{fig:sunburst} shows that the \textit{\textbf{Niger‑Congo}} and \textit{\textbf{Afro‑Asiatic}} families dominate the available resources. Within the \textit{\textbf{Niger‑Congo}} group, \textit{Kinyarwanda}, \textit{Yoruba}, \textit{Igbo}, \textit{Swahili}, and several \textit{Volta-Congo} languages account for the largest volumes of data. From the \textit{\textbf{Afro‑Asiatic}} family, \textit{Hausa} possesses substantial resources, whereas \textit{Somali}, \textit{Amharic}, and \textit{Tamazight} remain comparatively under‑represented. Other families like \textit{\textbf{Nilo‑Saharan}}, \textit{\textbf{Austronesian}}, and \textit{\textbf{Trans‑New Guinea}} appear only sparsely, with \textit{Malagasy} as the primary exception within the \textit{\textbf{Austronesian}} family. Overall, the majority of languages are from the \textbf{\textit{Niger-Congo}} and \textbf{\textit{Afro-Asiatic}} families, reflecting the dominant language groups in Africa.


\paragraph{Native Speaker Distribution.} We find a clear mismatch between speaker population size and available audio resources. Languages with large speaker populations often have minimal data—for example, \textit{Oromo}, with $45$M speakers, has only $34$ hours of audio, while \textit{Nigerian Pidgin}, spoken by roughly $120$M people, has just $0.21$ hours. In contrast, some languages with smaller populations are comparatively well‑resourced, such as \textit{South Ndebele} ($2.4$M speakers, $223$ hours) and \textit{Swati} ($4.7$M speakers, $307$ hours). These disparities suggest that data availability correlates more strongly with a number of potential factors such as language use in media, data archiving and accessibility, and targeted collection initiatives than with population size. Collectively, these factors are directly related to adopted language policies~\cite{adebara2025sahara}. 

\noindent \textbf{Number of Sources.} The number of data sources per language indicates that overall volume is primarily driven by inclusion in major collection projects rather than by a broad diversity of smaller efforts. Figure~\ref{fig:distrib_analysis_c} illustrates this relationship between source diversity and total hours collected. High‑volume languages either appear in multiple major sources—such as \textit{Hausa} (7 sources) and \textit{Swahili} (6 sources)—or derive substantial coverage from a single extensive initiative, as with \textit{Kinyarwanda} via CV-19 and \textit{Igbo} via NaijaVoice. In contrast, lower‑resource languages are typically represented only through isolated small‑scale efforts. Overall, data volume is dictated more by the scale of one or two dominant collections than by the sheer number of sources. A single large dataset can secure extensive hours but often limited in domain diversity, whereas multiple smaller sources may yield less total audio yet provide broader, more balanced coverage for downstream speech applications.

\section{\ourbenchmark Benchmark}\label{subsec:data_simbabench}
\noindent \textbf{Motivation.} To address the lack of standardized benchmarks for African speech technologies, we introduce \ourbenchmark—a unified evaluation suite designed to support diverse African speech tasks. It enables consistent model assessment, fosters reproducible research, and promotes fair comparisons, advancing inclusive language technologies for underrepresented communities.

\noindent \textbf{Coverage.} \ourbenchmark unifies all publicly available African speech datasets (Section~\ref{sec:data}), encompassing a wide range of languages, dialects, and domains. It supports comprehensive evaluation across both high- and low-resource languages through three core tasks: ASR, TTS, and SLID. Each task is paired with curated datasets and standardized metrics to enable consistent, fair comparisons across models and languages.



\paragraph{Data Splits and Release.} 
To ensure consistency and reproducibility, we adopt official training and test splits when available; otherwise, we apply a 90\%-10\% train-test partition. For model development and checkpoint selection, we construct a multilingual training and development set by sampling \textit{n} hours of training data (5 hours per language for ASR, 12 for TTS) and 30 minutes of development data per language when available. Evaluation is conducted per dataset to enable comparability with prior work and highlight dataset-specific challenges. We release the multilingual training and development splits to support benchmarking and tuning, while test sets are shared via standardized configuration files. \ourbenchmark will be hosted on the Hugging Face Datasets platform.

\section{Model Evaluation on \ourbenchmark}\label{sec:eval_othersmodel_evaluation}

\begin{table*}[!ht]
\centering
\resizebox{0.99\textwidth}{!}{%
\begin{tabular}{llrrrrrrrrrrH}
\toprule
\multicolumn{1}{c}{} & \multicolumn{1}{c}{} & &&& & \multicolumn{5}{c}{\textbf{\ourmodel Series (\textcolor{red}{Ours})}} & \multicolumn{1}{c}{} \\
 \cmidrule{8-12}
\multicolumn{1}{c}{\multirow{-2}{*}{\textbf{Language}}} & 
\multicolumn{1}{c}{\multirow{-2}{*}{\textbf{Test Set}}} & 
\multirow{-2}{*}{\textbf{MMS}} & \multirow{-2}{*}{\textbf{Seamless}} & \multirow{-2}{*}{\textbf{Whisper}} & \multirow{-2}{*}{\textbf{WhisperT}} & & 
\textbf{\ourmodel-H} & \textbf{\ourmodel-M} & \textbf{\ourmodel-S} & \textbf{\ourmodel-X} & \textbf{\ourmodel-W} & 
\multicolumn{1}{H}{\multirow{-2}{*}{\textbf{\ourmodel}}} \\
\midrule

Akuapim-twi (aka) &FS&85.82/40.14&\linecolor{red!100}{219.67/190.49}&\linecolor{red!100}{1181.0/1131.23}&\linecolor{red!100}{499.51/547.24}&&26.83/10.13&17.6/8.13&\colorbox{green!20}{13.29/8.45}&23.74/10.35&29.1/19.1 & 36.64/22.64\\ \midrule
Asante-twi (aka)&FS&83.6/32.35&\linecolor{red!100}{230.88/196.71}&\linecolor{red!100}{665.34/574.27}&\linecolor{red!100}{245.5/222.37}&&26.78/7.36&13.87/5.38&\colorbox{green!20}{7.06/2.62}&19.93/7.06&15.63/7.98& 27.09/14.44\\ \midrule
Afrikaans (afr)&Lwazi&92.06/37.59&37.91/16.47&66.05/34.32&73.17/39.05&&62.81/17.9&36.29/9.86&\colorbox{green!20}{15.62/4.99}&102.96/53.45&29.22/11.0&38.09/12.46\\
 ... & ... & ... & ... & ... & ... & &  ... & ... & ... & ... & ... & ... \\ 
  ... & ... & ... & ... & ... & ... & &  ... & ... & ... & ... & ... & ... \\ 
 Zulu (zul)&Lwazi&70.12/32.66&107.96/84.77&164.54/106.64&78.11/43.35&&62.92/17.57&38.58/10.88&108.53/103.61&101.93/52.87&\colorbox{green!20}{27.63/10.87}&\\
 Zulu (zul)&NCHTL&31.31/5.12&74.28/20.56&648.45/244.13&379.87/134.73&&30.55/4.69&26.36/3.96&\colorbox{green!20}{23.87/4.47}&60.96/8.79&33.92/5.71&\\ \midrule

\multicolumn{2}{c}{\textbf{Overall Average}} &75.9/35.26&146.69/98.92&611.91/437.98&196.7/149.79&&59.9/21.46&48.11/17.41&\colorbox{green!20}{41.65/18.3}&82.64/39.31&60.56/31.16&\\ \midrule

\end{tabular}%
}
\caption{Comparison of ASR performance across various African languages using baseline models and our \ourmodel models. Evaluation metrics are reported as \texttt{WER/CER}. \linecolor{red}{Red underlines} indicate that the model does not support the corresponding language, while \colorbox{green!20}{green highlights} denote the best-performing model for each language or test set. Full results are provided in Table~\ref{appdx_table:asr_results_full} (Appendix~\S\ref{appdx_sec:results}).}

\label{fig:asr_results} 
\end{table*}

We evaluate \ourbenchmark on several leading \textit{open}-source models to asses their generalization ability in the contexts of African languages and provide insights for future model development. Below we describe our evaluation pipeline in detail and the baseline models used for evaluation.

\begin{figure}[t]
  \centering
  \includegraphics[width=0.8\columnwidth]{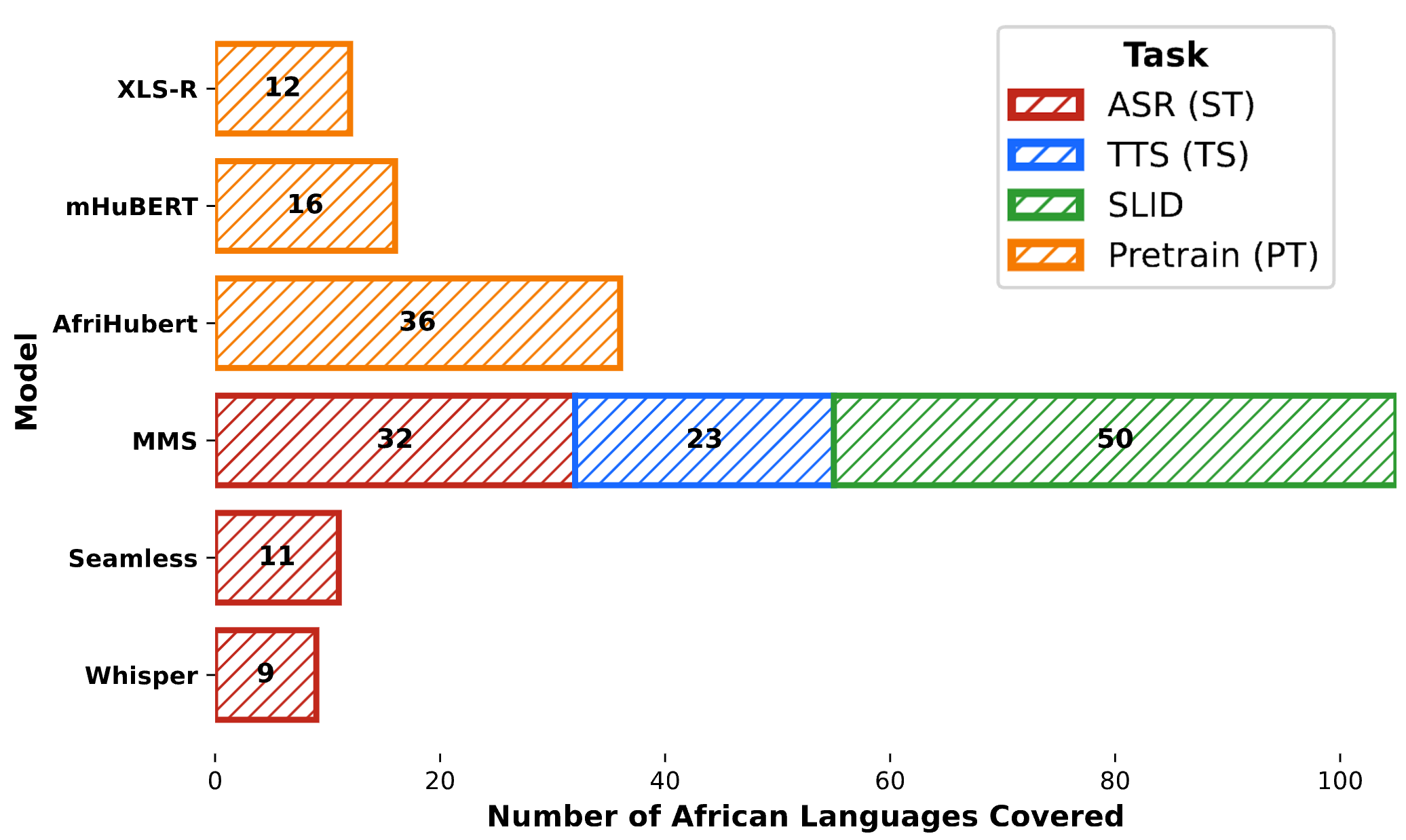}
\caption{Comparison of African language coverage across the downstream tasks as well as pretraining. }

\label{fig:african_coverage_by_model} 
\end{figure}

\subsection{Baseline Models}\label{subsec:baseline_models}
We benchmark several state-of-the-art multilingual speech models with varying architectures and training approaches to assess their performance on African language audio data. We evaluate Whisper~\cite{whisper}, Seamless~\cite{seamlessm4t2023}, MMS~\cite{pratap2023mms}, AfriHUBERT~\cite{alabi2024afrihubert}, and Wav2Vec2-XLS-R~\cite{babu2021xls}.

Figure~\ref{fig:african_coverage_by_model} illustrates the extent to which these baseline models cover African languages in their pretraining or supervised finetuning. The figure shows that MMS offers the broadest African language coverage across tasks, while models like AfriHUBERT provide the highest coverage in unsupervised pretraining. Whisper-v3 and SeamlessM4T-v2 provide limited ASR support, highlighting both task-specific strengths and existing gaps in African language inclusion. Table~\ref{appdx_tab:model_support_types}(Appendix \S\ref{appdx_sec:baseline_models}) presents a detailed overview of African language support across models for pretraining and various downstream tasks in speech and language processing.
Collectively, these models establish strong baselines for evaluating the current state of ASR technology for African languages.




\subsection{\ourmodel Series}\label{subsec:evaluation_simba}
In addition to evaluating existing speech models as described above, we finetune a series of models, referred to as the \ourmodel Series, leveraging the multilingual training and development sets from \ourbenchmark for the three downstream tasks. The \ourmodel models are designed to enhance performance and mitigate language coverage gaps identified in prior baselines.

\paragraph{\ourmodel-ASR.} We finetune five baseline models (see~\S\ref{subsec:baseline_models} for details) using the \ourbenchmark multilingual training and development sets.\footnote{ASR finetune data Comprise 215 hours of transcribed training audio (5 hours per language) and 21.5 hours of validation audio (30 minutes per language), covering 43 African languages.}  This multilingual setup enables the development of five new ASR models, each adapted specifically to African linguistic contexts. The resulting models are \ourmodel-H, finetuned from AfriHuBERT, \ourmodel-M from MMS-1b-all, \ourmodel-S from SeamlessM4T-v2-MT, \ourmodel-X from Wav2Vec2-XLS-R, and \ourmodel-W from Whisper-v3-large. All models are finetuned in a multilingual fashion. We follow the same protocols for multilingual training as described in the original Whisper, MMS, and Seamless models. For XLS-R, mHuBERT, and AfriHubert, we adopt a simple strategy of multilingual finetuning by adding a CTC layer and updating all parameters.  

\paragraph{\ourmodel-TTS.}
As the only baseline model that supports TTS, we finetune the MMS-TTS model~\cite{pratap2023mms} extending support to additional African languages. The original MMS-TTS model only supports $4$ out of the $11$ African languages included in our collection. As a result, unlike the ASR setup, we do not finetune on the entire multilingual dataset; instead, we focus exclusively on the $7$ African languages previously not supported by MMS-TTS, and for which TTS data exist in our collection. For each language, we independently finetune from existing MMS-TTS checkpoints belonging to linguistically similar languages, selecting the best-performing checkpoint based on validation performance. Specifically, \textit{Akuapem Twi} and \textit{Asante Twi} are finetuned from the \textit{Akan} checkpoint; \textit{Tswana} and \textit{Southern Sotho} from the \textit{Tsonga} checkpoint; \textit{Afrikaans} from the \textit{Dutch-based creole }checkpoint, reflecting its linguistic history; and \textit{Lingala} from the \textit{Swahili} checkpoint. This language-family-based knowledge transfer facilitates effective adaptation for these low-resource African languages.

\paragraph{\ourmodel-SLID.} Following the same ASR setup, we finetune AfriHuBERT, the pretrained model with the broadest African language coverage, using the 215-hour multilingual training split from \ourbenchmark. We validate on the corresponding 21.5-hour development set. This multilingual adaptation supports robust cross-lingual generalization for spoken language identification across diverse African languages.

\begin{table}[!ht]
\centering
\resizebox{0.80\columnwidth}{!}{%
\begin{tabular}{lllrH}
\toprule
\textbf{Setting}           & \textbf{Langauge}         & \textbf{Test Set} & \multicolumn{1}{l}{\textbf{WER/CER}} & \multicolumn{1}{H}{\textbf{Our}} \\ \midrule
\multirow{5}{*}{\rotatebox{90}{\textbf{MMS-TTS}}} & Ewe (ewe)                 & bibleTTS          & 15.94/3.42                            &                              \\
                           & Yoruba (yor)              & bibleTTS          & 26.99/9.66                            &                             \\
                           & Hausa (hau)               & bibleTTS          & 14.09/3.15                           &                              \\
                           & Kinyarwanda (kin)         & KinyarwandaTTS    & 44.75/8.95         &          \\ \cmidrule{2-5}
                           \multicolumn{3}{c}{\textbf{ Average}}& 25.44/6.30          &           \\ \midrule
\multirow{6}{*}{\rotatebox{90}{\textbf{\ourmodel-TTS} }}  & \linecolor{red!100}{Xhosa (xho)}               & SouthAfricaTTS    & 71.98/23.34                             &                            \\
                           & \linecolor{red!100}{Lingala (lin)}            & bibleTTS          & 35.97/6.94                            &                              \\
                           & \linecolor{red!100}{Asante-twi (aka)}   & bibleTTS          & 59.84/19.32                             &                            \\
                           & \linecolor{red!100}{Akuapim-twi (aka)} & bibleTTS          & 59.45/18.29                             &                            \\
                           & \linecolor{red!100}{Afrikaans (afr) }          & SouthAfricaTTS    & 78.31/35.02          &     \\ 
                           & \linecolor{red!100}{Tswana (tsn) }          & SouthAfricaTTS    & 90.30/44.93          &     \\ 
                           & \linecolor{red!100}{Southern Sotho (sot) }          & SouthAfricaTTS    & 91.84/44.20          &     \\ 
                           \cmidrule{2-5}
                           \multicolumn{3}{c}{\textbf{Average}}& 68.23/26.17         &           \\ \midrule
                           \multicolumn{3}{l}{\textbf{Overall Average}}& 46.84/16.23        &           \\ 
                           
                           \bottomrule
\end{tabular}%
}
\caption{Performance of the Original MMS-TTS on Supported Languages and finetuned \ourmodel-TTS on Unsupported Languages. \linecolor{red!100}{Red Underline} indicates languages that are not supported by the MMS-TTS model.}
\label{tab:tts-results}
\end{table}
\subsection{Evaluation Pipeline}\label{subsec:evaluation_strategy}
Our evaluation pipeline is designed to ensure consistency across downstream tasks and models, providing a robust framework for analyzing performance under varying resource constraints. As shown in Figure~\ref{fig:data_workflow}, our evaluation pipeline relies on two settings: (i) zero-shot evaluation of baseline models\footnote{These models are already fine-tuned on task-specific data; however, we refer to this as zero-shot since we evaluate them on languages that are unsupported or unseen during training.}, specifically targeting languages not seen during training or not officially supported; and (ii) evaluation of finetuned models to quantify adaptation gains. Detailed information about the experimental setup, hyperparameters, and evaluation metrics is provided in Appendix \ref{appdx_sec:hyparam}.



\section{Results}\label{sec:result}

Table~\ref{fig:asr_results} presents the performance of baseline systems and our \ourmodel-ASR models on \ourbenchmark across $56$ language-specific test sets representing $46$ languages on the ASR task. Among the $23$ test sets for which none of the baseline models officially support, MMS achieves the best performance across all baselines. This trend is particularly evident for languages such as \textit{Standard Moroccan Tamazight}, \textit{Venda}, \textit{Tswana}, \textit{Swati}, \textit{Sotho}, and \textit{Northern Ndebele}. However, several languages remain challenging for all evaluated models. Specifically, \textit{Susu}, \textit{Tigre}, \textit{Tigrinya}, and \textit{Ga} consistently yield high error rates, revealing substantial gaps in support for certain under-resourced languages. Our finetuned \ourmodel-ASR models improve upon every test sets compared to the baseline systems, with \ourmodel-S achieving the best overall performance, reaching $41.65$ WER and $18.30$ CER. These improvements underscore the effectiveness of model adaptation for African languages, with significant improvements for several previously unsupported languages like \textit{Fanti}, \textit{Venda}, \textit{Swati}, and \textit{Bemba}. However, certain languages—including \textit{Western Maninkakan}, \textit{Tigrinya}, \textit{Standard Moroccan Tamazight}, and \textit{Susu}—continue to exhibit high error rates (exceeding $100$ WER), indicating that further progress will require additional data and more targeted modeling strategies.

Table~\ref{tab:tts-results} presents results on the TTS task across both supported and unsupported languages. MMS-TTS demonstrates relatively strong performance on the languages it officially supports, with  low error rates for \textit{Hausa} ($14.09$/$3.15$) and \textit{Ewe} ($15.94$/$3.42$). Performance remains competitive  for \textit{Yoruba} ($26.99$/$9.66$), but drops significantly for \textit{Kinyarwanda} ($44.75$/$8.95$), highlighting that even among supported languages, synthesis quality can vary considerably. Our finetuned \ourmodel-TTS models despite limited training data, achieve reasonable results by finetuning for each target language. Nevertheless, error rates remain high: $78.31$/$35.02$ for \textit{Afrikaans}, $71.98$/$23.34$ for \textit{Xhosa}, $59.84$/$19.32$ for \textit{Asante-Twi}, and $59.45$/$18.29$ for \textit{Akuapem-Twi}. Interestingly, we observe improved performance on data derived from BibleTTS, likely due to the domain’s relatively constrained linguistic structure and vocabulary, which appear to support more consistent synthesis.

Table~\ref{appdx_tab:slid-results} reports SLID performance across 32 language-dataset pairs using MMS-LID-1024 and \ourmodel-SLID. While MMS performs well on high-resource languages, \ourmodel-SLID shows notable gains on low-resource languages, addressing key identification gaps.
\section{Discussions}\label{sec:discussion}
\paragraph{Dataset Variation.}
We find that dataset variations strongly impact performance. On the ASR task, MMS and Seamless models show significantly better performance on \textit{Afrikaans} CV-$19$ compared to Lwazi and NCHLT datasets. Additionally, both \textit{Zulu} and \textit{Xhosa} consistently achieve better performance on NCHLT datasets than on the Lwazi datasets. This performance gap likely stems from dataset quality differences: NCHLT features broadband speech recordings with over $50$ hours per language, while Lwazi contains telephone speech recordings with only $4$-$10$ hours per language, providing more diverse, higher-quality training material in NCHLT. On the SLID task, \textit{Hausa} scores $100$\% on OlongoAfrica but only $75$\% on UDHR. This discrepancy likely stems from domain differences: UDHR contains human rights declarations with specialized vocabulary that might complicate language identification, while OlongoAfrica features short stories with more natural language patterns that preserve distinctive linguistic features, making identification easier. Similarly, on the TTS task, test sets drawn from the Bible domain consistently yield lower error rates than those from other domains such as KinyarwandaTTS or SouthAfricaTTS, underscoring the strong influence of domain characteristics on model performance. This reinforces the need for diverse, representative test sets when evaluating multilingual models.

\paragraph{Model Task Coverage.} Notably, sheer language coverage does not guarantee uniformly strong ASR accuracy. MMS, which supports the largest number of African languages, attains the best overall average, confirming that extensive pre-training across many languages yields broad, reliable results. Yet this advantage does not extend to every high-resource language: for \textit{Amharic} and \textit{Afrikaans}, Seamless—with far smaller coverage—occasionally surpasses MMS, suggesting that focused training and larger model size can overcome limited coverage when sufficient in-domain data exist. Conversely, Whisper, covering only nine African languages, records the highest error rates overall, and its performance collapses for the many languages it does not officially support, underscoring how lack of task-specific training degrades performance. Overall, wide coverage prevents failure on unsupported languages, whereas fine-grained adaptation determines which system performs best among languages that are already supported.

\paragraph{Relation to Language Family.} Our analysis reveals that language family relationships significantly influence model performance patterns across tasks. Within the \textbf{\textit{Niger-Congo}} family, closely related \textit{Volta-Congo} languages like \textit{Swahili}, \textit{Zulu}, and \textit{Xhosa} demonstrate similar performances, particularly the \ourmodel series models. Low-resource languages benefit substantially from relationships in well-represented families, languages from the \textit{Mande} group achieve reasonable performance despite limited training data, likely due to transfer learning from related \textbf{\textit{Niger-Congo}} languages. The effect is especially apparent for \textbf{\textit{Afro-Asiatic}} languages; \textit{Amharic} performs exceptionally well with \ourmodel-X
despite moderate training data, suggesting effective cross-lingual knowledge transfer within its family. These patterns indicate that models leverage shared linguistic features within families, confirming that while comprehensive family representation in training data significantly impacts potential performance, the strength of family representation in training data significantly impacts potential performance, especially for lower-resourced languages of well-represented families.

\section{Conclusion}\label{sec:conc}
In this work, we present \ourbenchmark, a large-scale benchmark covering $61$ African languages across core speech downstream tasks. By curating over $8,600$ hours of speech data from diverse domains  and language families, we enable comprehensive evaluation of multilingual and Africa-centered speech models. Using \ourbenchmark, we finetune the \ourmodel series—task-specific models that enhance performance and mitigate language coverage gaps identified in prior baselines, achieving SoTA results on many low-resource languages. We find that, while broad language coverage provides a useful baseline, our analysis shows that model performance is strongly influenced by domain diversity, data quality, and linguistic relatedness. Our findings underscore the importance of multilingual adaptation and language-family-aware training, highlighting \ourbenchmark as a critical tool for advancing inclusive African speech technologies.

\section{Limitations}\label{sec:limits}
Our study has a number of limitations that highlight important avenues for future work:

\begin{enumerate}
    \item \textbf{Data Availability and Representation Bias.} \ourbenchmark relies solely on publicly available datasets, which reflect existing structural and historical biases in language technology development. Many Indigenous African languages remain severely underrepresented, limiting the benchmark’s ability to capture the full spectrum of Africa’s linguistic diversity.

    \item \textbf{Task Coverage.} Our evaluation is restricted to three core ASR, TTS, and SLID due to data availability. Broader downstream tasks such as speech translation, spoken question answering, or spoken dialogue systems are not yet supported and require further dataset development.

    \item \textbf{Modeling Scope.} We focus on finetuning existing models rather than proposing new architectural innovations or advanced adaptation methods. While our results demonstrate the benefits of task-specific tuning, we do not explore complementary strategies such as self-supervised pretraining, multitask learning, or data augmentation.

    \item \textbf{Implementation Constraints.} Despite our advocacy for inclusive data and policy reform, real-world implementation requires sustained institutional commitment. Bridging the gap between research and impact will necessitate long-term investment from governments, academia, and industry partners.

    \item \textbf{Task Diversity and Generalization.} Although \ourbenchmark spans three speech tasks, it does not yet cover interactive or generative applications such as conversational AI, spoken retrieval, or end-to-end multilingual agents. Extending the benchmark to include such tasks would further promote holistic model evaluation and real-world applicability.
\end{enumerate}

Despite these limitations, our work emphasizes the urgency of addressing speech data disparities and fostering inclusive language technologies across the African continent.


\section{Ethical Considerations}\label{sec:ethics}
We outline several ethical considerations relevant to this work:

\begin{enumerate}
    \item Our research aims to advance speech technology for African languages by addressing the historical marginalization of many linguistic communities and promoting equitable digital inclusion across the continent.

    \item The datasets used in our benchmark are sourced from publicly available repositories. However, their existence reflects broader sociopolitical dynamics, including which languages have received institutional support and technological investment. This highlights the role of policy in shaping digital language presence.

    \item Although we do not propose novel model architectures, we fine-tune existing models on \ourbenchmark and release stronger task-specific checkpoints. Our analysis illustrates how unequal data availability—shaped by historical and policy-driven neglect—affects performance, underscoring the need for targeted policy interventions to support multilingual data creation and ethical development.

    \item We stress the importance of proper attribution for both datasets and models, as a matter of transparency, accountability, and fair recognition. To this end, we provide a publicly accessible reference list citing all datasets and fine-tuned models used in our benchmark, and encourage researchers and institutions to uphold responsible and inclusive data stewardship.
\end{enumerate}

\section*{Acknowledgments}\label{sec:acknow}
We acknowledge support from Canada Research Chairs (CRC), CLEAR Global for funding from the Gates Foundation, the Natural Sciences and Engineering Research Council of Canada (NSERC; RGPIN-2018-04267), the Social Sciences and Humanities Research Council of Canada (SSHRC; 895-2020-1004; 895-2021-1008), Canadian Foundation for Innovation (CFI; 37771), Digital Research Alliance of Canada,\footnote{\href{https://alliancecan.ca}{https://alliancecan.ca}} and UBC ARC-Sockeye.\footnote{\href{https://arc.ubc.ca/ubc-arc-sockeye}{https://arc.ubc.ca/ubc-arc-sockeye}} We also thank Hellina Hailu Nigatu, Atnafu Lambebo, and Wei-Rui Chen for discussions related to this work. The findings and conclusions
contained within this work those of the authors and do not necessarily reflect positions or policies of any supporters.
\normalem
\bibliography{custom}
\appendix
\clearpage
\appendix
\appendixpage            
\addappheadtotoc         
\numberwithin{figure}{section}
\numberwithin{table}{section}

The following appendices provide comprehensive supplementary material supporting the main findings of this work. We include detailed descriptions of the datasets used, data preprocessing steps, baseline models, experimental setup, and full evaluation results across tasks and languages. This material is intended to enhance reproducibility, offer deeper insight into model behavior, and serve as a resource for future research in African speech technologies. The appendices are organized as follows:

\begin{itemize}
    \item \S\ref{app_sec:data}: Data Collection and Corpus Curation
    \item \S\ref{appdx_sec:mapping}: Mapping the Data Landscape
    \item \S\ref{app_sec:preprocessing_pipeline}: Preprocessing Pipeline
    \item \S\ref{appdx_sec:baseline_models}: Baseline Models
    \item \S\ref{appdx_sec:hyparam}: Experimental Setup
    \item \S\ref{appdx_sec:results}: Evaluation Results
\end{itemize}

\textbf{Key tables include:
}
\begin{itemize}
    \item Table~\ref{appdex_tab:iso_codes_table}: Total duration of audio (in hours) available per language across multiple datasets.
    \item Table~\ref{appdx_tab:model_support_types}: Overview of African language coverage across models for pretraining and downstream speech and language tasks.
    \item Table~\ref{appdx_table:asr_results_full}: Comparison of ASR performance across various African languages using baseline models and our Simba models in both zero-shot and fine-tuned settings.
    \item Table~\ref{appdx_tab:slid-results}: Performance of MMS-LID-1024 and Simba-SLID on SimbaBench.
\end{itemize}

\section{Data Collection and Corpus Curation}\label{app_sec:data}
\subsection{Automatic Speech Recognition Data}

\paragraph{ALFFA PUBLIC Dataset} \cite{ALFFA_PUBLIC}: is a multilingual dataset developed as part of the ALFFA (African Languages in the Field: Speech Fundamentals and Automation) project. It supports ASR systems for under-resourced Sub-Saharan African languages and includes resources for Wolof ($5.74$ hours), Fongbe ($2.89$ hours), Amharic ($3.12$ hours), and Swahili ($3.93$ hours). 

\paragraph{Bemba Speech Dataset} \cite{sikasote-anastasopoulos:2022:LREC} consists of read speech compiled from various publicly available Bemba sources, including books, show transcripts, and YouTube transcripts. It contains 15,000 utterances totaling 24.5 hours of audio, making it a valuable resource for ASR and linguistic research for the Bemba language. 

\paragraph{Mozilla Common Voice} \cite{commonvoice} is a multilingual dataset designed to improve voice technologies for under-resourced languages. The African language collection includes significant contributions in a variety of languages, with notable amounts of recorded hours in Kinyarwanda (1,354.02 hours), Kabyle (174.66 hours), Ganda (149.43 hours), and Swahili (106.89 hours). Additional contributions include Kalenjin (29.88 hours), Luo (13.63 hours), Hausa (3.77 hours), Taita (4.47 hours), and smaller datasets for languages like Amharic (1.58 hours), Basaa (2.19 hours), and Standard Moroccan Tamazight (1.07 hours). This dataset provides a valuable resource for ASR systems and other linguistic technologies aimed at African languages. More information is available on Common Voice's official page\footnote{\href{https://commonvoice.mozilla.org}{https://commonvoice.mozilla.org}}.

\paragraph{Financial Inclusion Speech Dataset} \cite{FinancialInclusionSpeechDataset} is a multilingual speech dataset developed to support financial inclusion in Ghana. Created by Ashesi University and Nokwary Technologies, the dataset comprises recordings from approximately 200 speakers per language, each recording around 130 sentences. The languages covered include Akuapem Twi (38 hours), Asanti Twi (30 hours), Fanti (39 hours), and Ga (40 hours), totaling approximately 148 hours of speech data. 

\paragraph{Kallaama} \cite{Gauthier2024Kallaama} the Kallaama dataset is a rich resource of transcribed agricultural speech in Senegal's three most widely spoken languages: Wolof, Pulaar, and Sereer. Comprising more than 100 hours of spontaneous audio recordings from farmers, agricultural advisers, and agribusiness managers, the data include radio programs, focus groups, voice messages, and interviews.

\paragraph{Lwazi Speech Corpus} \cite{vanHeerden2016LwaziASR} is a multilingual dataset that includes telephone speech recordings in the 11 official languages of South Africa. Each language has approximately 200 speakers, each speaker reading an average of 30 prompts, resulting in 4 to 10 hours of audio per language. 


\paragraph{NaijaVoices Dataset} \cite{NaijaVoices2024} is a multilingual speech corpus designed to support ASR and NLP tasks in Nigerian languages. It includes approximately 1,800 hours of speech data and curated text in Yoruba, Igbo, and Hausa, with roughly 600 hours dedicated to each language.

\paragraph{NCHLT Speech Corpus} \cite{Barnard2014NCHLT} is a multilingual dataset of broadband speech collected from approximately 200 speakers per language in each of the 11 official languages of South Africa: Afrikaans, English, Ndebele, Northern Sotho, Southern Sotho, Swati, Tswana, Tsonga, Venda, Xhosa, and Zulu. Developed under the National Center for Human Language Technology (NCHLT) initiative, the corpus comprises more than 50 hours of orthographically transcribed speech for each language.

\paragraph{Nicolingua - West African Virtual Assistant Speech Recognition Corpus} \cite{doumbouya2021usingradio} is a multilingual dataset comprising 10,083 recorded utterances in four languages: Susu (51 hours), Western Maninkakan (42 hours), Pular (31 hours), and French. Collected from 49 speakers, the corpus is designed to support the development of speech recognition systems for West African languages. 

\paragraph{Yoruba Speech Dataset} \cite{gutkin-et-al-yoruba2020} is a high-quality crowdsourced dataset of Yoruba audio recordings designed for speech processing applications. It includes transcribed WAV files, with separate archives for female and male speakers and the corresponding transcription. It is manually quality-checked and provides valuable resources for developing ASR systems and other linguistic tools for Yoruba.

\paragraph{Zambezi Voice Project} \cite{sikasote23_interspeech} led by the University of Zambia speech and language research group, this ongoing initiative aims to create speech and language resources for Zambia's under-resourced native languages. The labeled dataset comprises more than 36,000 read-speech recordings totaling 79 hours, with contributions from Bemba (26 hours), Nyanja (25 hours), Tonga (22 hours), and Lozi (6 hours). The unlabeled dataset, derived from radio broadcasts, provides 525 hours of audio, including Bemba (162 hours), Tonga (101 hours), Lozi (30 hours), Nyanja (25 hours), and Lunda (39 hours). These resources support the development of ASR and other language technologies for Zambian languages.

\subsection{Text-To-Speech Data}

\paragraph{BibleTTS} \cite{meyer2022bibletts}: BibleTTS is a high-quality multilingual TTS corpus featuring up to 80 hours of studio-quality recordings for each of six Sub-Saharan African languages: Asante Twi, Akuapem Twi, Ewe, Hausa, Lingala, and Yoruba. Derived from the Biblica open.bible project, the dataset includes verse-aligned and filtered speech-text pairs. 


\paragraph{High quality TTS data for four South African Languages} \cite{van-niekerk-etal-2017}: Collected in collaboration between North-West University and Google, this dataset provides over 3 hours of high-quality, multi-speaker transcribed audio recordings for each of the four South African languages: Afrikaans, Sesotho, Setswana, and isiXhosa. 


\paragraph{Kinyarwanda TTS} \cite{DigitalUmuganda_afrispeak_2023}: is a high-quality Text-to-Speech corpus developed and hosted by Digital Umuganda on Hugging Face. The combined dataset totals approximately 14 hours of speech data, covering diverse phonetic contexts and speaking styles. 

\subsection{Spoken Language Identification Data}

\paragraph{NicoLingua - West African Radio Corpus} \cite{doumbouya2021usingradio}: This dataset contains 17,090 audio clips, each 30 seconds long, sampled from archives of Guinean radio stations. It spans 10 languages—French, Guerze, Koniaka, Kissi, Kono, Maninka, Mano, Pular, Susu, and Toma—totaling approximately 143.76 hours of audio. The recordings feature a variety of content, including news and radio shows, with rich acoustic diversity such as phone calls, background music, and environmental noise. A validation set of 300 manually tagged clips is included to support evaluation.

\paragraph{VoxLingua107 Dataset} \cite{Valk2021VoxLingua107}: is a large-scale multilingual spoken language recognition (SLR) corpus containing over 4,000 hours of YouTube speech data, automatically labeled using language-specific queries. It covers 107 languages and is freely available for research. The dataset includes African languages such as Swahili ($57.48$h), Somali ($92.47$h), Shona ($27.19$h), Amharic ($73.36$h), Hausa ($83.80$h), Yoruba ($84.66$h), Lingala ($81.31$h), Afrikaans ($97.46$h), and Malagasy ($98.27$h). In addition, we specifically selected high-resource non-African languages including Italian ($45.91$h), Portuguese ($58.03$h), Spanish ($34.95$h), Arabic ($52.88$h), and English ($43.84$h).


\subsection{New Raw Audio Data}

\paragraph{OlongoAfrica Multilingual Anthology} \cite{OlongoAfrica2024}: is a collection of translated and narrated short stories in 10 African languages, showcasing the linguistic diversity of the continent. The included languages are Edo, Tamazight, Yoruba, Swahili, Hausa, Tiv, Shona, Ibibio, Igbo, and Nigerian Pidgin.

\paragraph{UDHR} \cite{UDHRAudio2025}: The website UDHR.audio hosts raw audio recordings of the Universal Declaration of Human Rights (UDHR) in numerous languages. These recordings capture the text being read aloud by native speakers, Among the languages included, we specifically collected high-quality recordings for Hausa, Tem, Amharic, Wolof, Swahili, and Afrikaans.

\paragraph{VOA} \cite{VoiceOfAfrica2025}\footnote{\href{https://www.voaafrica.com/}{https://www.voaafrica.com/}}: Voice of Africa includes a collection of news websites delivering updates and stories from across the African continent. This dataset features meticulously collected news videos from the platform in languages such as Tigrinya, North Ndebele, Swahili, Oromo, Kinyarwanda, Somali, Hausa, Amharic, French, Shona, and Lingala, totaling over 1500 hours of speech content.

\subsection{Code-Switched Audio Data}

\paragraph{CS Soap Opera} \cite{VAN_DER_WESTHUIZEN18.90}: is a multilingual speech dataset compiled from South African soap operas, featuring code-switched speech between English and four Bantu languages: Zulu ($5.45$h), Xhosa ($3.13$h), Swana ($2.86$h), and Southern Sotho ($2.83$h). It includes multiple forms of code-switching, including between sentences, within sentences, and within individual words—making it a rich resource for studying multilingual ASR in the South African context.

\paragraph{SPCS} \cite{Modipa2015CodeSwitching}: is a 10.48-hour speech dataset featuring code-switched utterances between Sepedi and English. It was created to support ASR research on multilingual speech involving a minority Bantu language and captures natural switching patterns across diverse speakers and contexts.



\section{Mapping the Data Landscape}\label{appdx_sec:mapping}
Table~\ref{appdex_tab:iso_codes_table} provides detailed information on the total audio duration (in hours) available for each language across various datasets.

\begin{table*}[!ht]
\centering
\small
\renewcommand{\arraystretch}{0.99}
\resizebox{0.99\textwidth}{!}{%
\begin{tabular}{llr|p{0.65\textwidth}}
\toprule
\textbf{Language} & \textbf{ISO-3} & \textbf{Hours} & \textbf{Dataset Breakdown (Color-coded)} \\
\midrule
Afrikaans & afr & 255.3 & \dsentry{lwazicolor}{4.28} \dsentry{nchtlcolor}{138.73} \dsentry{sattcolor}{3.31} \dsentry{udhrcolor}{0.15} \dsentry{voxcolor}{108.39} \dsentry{cv19color}{0.43} \\
Akuapim-twi & aka & 98.92 & \dsentry{biblecolor}{60.57} \dsentry{finspeechcolor}{38.35} \\
Asante-twi & aka & 31.96 & \dsentry{biblecolor}{1.53} \dsentry{finspeechcolor}{30.43} \\
Amharic & amh & 107.78 & \dsentry{alffacolor}{20.76} \dsentry{udhrcolor}{0.05} \dsentry{voacolor}{3.91} \dsentry{voxcolor}{81.47} \dsentry{cv19color}{1.59} \\
Basaa & bas & 2.19 & \dsentry{cv19color}{2.19} \\
Bemba & bem & 92.81 & \dsentry{bembacolor}{26.93} \dsentry{zambezi2color}{65.88} \\
Taita & dav & 4.47 & \dsentry{cv19color}{4.47} \\
Dyula & dyu & 0.32 & \dsentry{cv19color}{0.32} \\
Edo & bin & 0.18 & \dsentry{ologoafricacolor}{0.18} \\
Ewe & ewe & 77.63 & \dsentry{biblecolor}{77.63} \\
Fanti & fat & 39.84 & \dsentry{finspeechcolor}{39.84} \\
Fon & fon & 7.18 & \dsentry{alffacolor}{7.18} \\
Pulaar & fuc & 24.8 & \dsentry{kallaamacolor}{24.8} \\
Pular & fuf & 0.52 & \dsentry{nicolingua3color}{0.21} \dsentry{nicolingua4color}{0.31} \\
Ga & gaa & 40.93 & \dsentry{finspeechcolor}{40.93} \\
Hausa & hau & 908.42 & \dsentry{naijacolor}{617.95} \dsentry{ologoafricacolor}{0.14} \dsentry{udhrcolor}{0.19} \dsentry{voacolor}{106.5} \dsentry{voxcolor}{93.3} \dsentry{biblecolor}{86.57} \dsentry{cv19color}{3.78} \\
Ibibio & ibb & 0.31 & \dsentry{ologoafricacolor}{0.31} \\
Igbo & ibo & 634.95 & \dsentry{naijacolor}{634.59} \dsentry{ologoafricacolor}{0.33} \dsentry{cv19color}{0.02} \\
Kabyle & kab & 174.66 & \dsentry{cv19color}{174.66} \\
Tem & kdh & 0.29 & \dsentry{udhrcolor}{0.29} \\
Kinyarwanda & kin & 1374.35 & \dsentry{kinyacolor}{14.08} \dsentry{voacolor}{6.25} \dsentry{cv19color}{1354.01} \\
Kalenjin & kln & 29.87 & \dsentry{cv19color}{29.87} \\
Guerze/Kpelle & kpe & 0.09 & \dsentry{nicolingua3color}{0.09} \\
Kisi & kss & 0.05 & \dsentry{nicolingua3color}{0.05} \\
Lingala & lin & 201.62 & \dsentry{voacolor}{56.1} \dsentry{voxcolor}{90.26} \dsentry{biblecolor}{55.26} \\
Lozi & loz & 21.64 & \dsentry{zambezi1color}{6.22} \dsentry{zambezi2color}{15.42} \\
Ganda & lug & 149.42 & \dsentry{cv19color}{149.43} \\
Lunda & lun & 20.47 & \dsentry{zambezi2color}{20.47} \\
Luo (Kenya and Tanzania) & luo & 13.62 & \dsentry{cv19color}{13.62} \\
Konyanka Maninka & mku & 0.12 & \dsentry{nicolingua3color}{0.12} \\
Malagasy & mlg & 109.21 & \dsentry{voxcolor}{109.21} \\
Western Maninkakan & mlq & 0.42 & \dsentry{nicolingua4color}{0.42} \\
Mandinka & mnk & 0.63 & \dsentry{nicolingua3color}{0.63} \\
South Ndebele & nbl & 223.88 & \dsentry{lwazicolor}{4.28} \dsentry{nchtlcolor}{219.6} \\
North Ndebele & nde & 14.05 & \dsentry{voacolor}{14.05} \\
Northern Sotho (Sepedi) & nso, eng-nso & 188.43 & \dsentry{lwazicolor}{4.28} \dsentry{nchtlcolor}{173.66} \dsentry{cv19color}{0.0} \dsentrycs{spcscolor}{10.48}{CS - English} \\
Nyanja & nya & 36.51 & \dsentry{zambezi1color}{25.34} \dsentry{zambezi2color}{11.17} \\
Oromo & orm & 34.55 & \dsentry{voacolor}{34.55} \\
Nigerian Pidgin & pcm & 0.21 & \dsentry{ologoafricacolor}{0.21} \\
Shona & sna & 39.55 & \dsentry{ologoafricacolor}{0.3} \dsentry{voacolor}{8.97} \dsentry{voxcolor}{30.29} \\
Somali & som & 192.12 & \dsentry{voacolor}{89.32} \dsentry{voxcolor}{102.8} \\
Southern Sotho & sot, eng-sot & 184.67 & \dsentry{lwazicolor}{4.28} \dsentry{nchtlcolor}{174.34} \dsentry{sattcolor}{3.22} \dsentrycs{soapcolor}{2.83}{CS - English} \\
Serer & srr & 34.38 & \dsentry{kallaamacolor}{34.38} \\
Susuami & ssu & 0.23 & \dsentry{nicolingua3color}{0.23} \\
Swati & ssw & 307.04 & \dsentry{lwazicolor}{4.28} \dsentry{nchtlcolor}{302.76} \\
Susu & sus & 0.51 & \dsentry{nicolingua4color}{0.51} \\
Swahili & swa & 689.27 & \dsentry{ologoafricacolor}{0.28} \dsentry{udhrcolor}{0.19} \dsentry{voacolor}{506.27} \dsentry{voxcolor}{63.89} \dsentry{cv19color}{106.89} \dsentry{alffacolor}{11.75} \\
Tigre & tig & 1.04 & \dsentry{cv19color}{1.04} \\
Tigrinya & tir & 39.2 & \dsentry{voacolor}{39.16} \dsentry{cv19color}{0.04} \\
Tiv & tiv & 0.27 & \dsentry{ologoafricacolor}{0.27} \\
Tonga (Zambia) & toi & 85.72 & \dsentry{zambezi1color}{22.67} \dsentry{zambezi2color}{63.06} \\
Tswana & tsn, eng-tsn & 174.7 & \dsentry{lwazicolor}{4.28} \dsentry{nchtlcolor}{164.03} \dsentry{sattcolor}{3.52} \dsentry{cv19color}{0.0} \dsentrycs{soapcolor}{2.86}{CS - English} \\
Tsonga & tso & 145.24 & \dsentry{lwazicolor}{4.28} \dsentry{nchtlcolor}{140.96} \\
Twi & twi & 0.21 & \dsentry{cv19color}{0.21} \\
Central Atlas Tamazight & tzm & 0.26 & \dsentry{ologoafricacolor}{0.26} \\
Venda & ven & 209.86 & \dsentry{lwazicolor}{4.28} \dsentry{nchtlcolor}{205.58} \\
Wolof & wol & 73.65 & \dsentry{alffacolor}{18.97} \dsentry{kallaamacolor}{54.5} \dsentry{udhrcolor}{0.18} \\
Xhosa & xho, eng-xho & 225.42 & \dsentry{lwazicolor}{4.28} \dsentry{nchtlcolor}{214.89} \dsentry{sattcolor}{3.11} \dsentry{cv19color}{0.0} \dsentrycs{soapcolor}{3.13}{CS - English} \\
Yoruba & yor & 738.31 & \dsentry{naijacolor}{614.98} \dsentry{ologoafricacolor}{0.12} \dsentry{voxcolor}{94.05} \dsentry{yourbavoicecolor}{4.03} \dsentry{biblecolor}{25.13} \\
Standard Moroccan Tamazight & zgh & 1.07 & \dsentry{cv19color}{1.07} \\
Zulu & zul, eng-zul & 197.24 & \dsentry{lwazicolor}{4.28} \dsentry{nchtlcolor}{187.5} \dsentry{cv19color}{0.01} \dsentrycs{soapcolor}{5.45}{CS - English} \\ \midrule

Multiple\textsuperscript{$\star$} &  & 142.42 & \dsentry{nicolingua3color}{142.42} \\
\midrule
English - Accented & eng & 200 & \dsentry{afrispeechcolor}{200} \\
\bottomrule
\end{tabular}
}

\caption{Total duration of audio (in hours) available per language across multiple datasets. Color-coded cells indicate the contributing datasets for each language. \textsuperscript{$\star$}The ``Multiple'' row refers to unlabeled audio data encompassing the following languages: \texttt{kpe}, \texttt{kss}, \texttt{mku}, \texttt{mnk}, \texttt{fuf}, and \texttt{ssu}.}

\captionsetup{labelformat=empty,font=footnotesize,justification=raggedright}
\caption*{%
  \parbox{\textwidth}{%
    \textbf{Dataset Legend:}\\[0.75ex]
    \raggedright
    \dsbox{afrispeechcolor} afrispeech-200 \quad
    \dsbox{alffacolor} Alffa\_Public \quad
    \dsbox{bembacolor} BembaSpeech \quad
    \dsbox{biblecolor} bibleTTS \quad
    \dsbox{cv19color} Common Voice 2019 \quad
    \dsbox{finspeechcolor} fin\_speech \quad
    \dsbox{kallaamacolor} Kallaama \quad
    \dsbox{kinyacolor} KinyarwandaTTS \quad
    \dsbox{lwazicolor} Lwazi \quad
    \dsbox{naijacolor} NaijaVoice \quad
    \dsbox{nchtlcolor} NCHTL / AUX \quad
    \dsbox{nicolingua3color} Nicolingua-0003 \quad
    \dsbox{nicolingua4color} Nicolingua-0004 \\[0.5ex]
    \dsbox{ologoafricacolor} OlogoAfrica \quad
    \dsbox{sattcolor} SouthAfricaTTS \quad
    \dsbox{soapcolor} CS\_Soap\_Opera \quad
    \dsbox{spcscolor} SPCS \quad
    \dsbox{udhrcolor} UDHR \quad
    \dsbox{voacolor} VOA \quad
    \dsbox{voxcolor} VoxLingua \quad
    \dsbox{yourbavoicecolor} YorubaVoice \\[0.5ex]
    \dsbox{zambezi1color} ZambeziVoice \quad
    \dsbox{zambezi2color} ZambeziVoice (ULB)
  }
}

\label{appdex_tab:iso_codes_table}

\end{table*}

\section{Preprocessing Pipeline}\label{app_sec:preprocessing_pipeline}
\paragraph{Audio Standardization.} All recordings were resampled to a uniform sampling rate of $16$~kHz and converted to single-channel (mono) WAV format. This step ensures compatibility across toolkits and mitigates discrepancies caused by varying source formats and encodings.

\paragraph{Segmentation, Filtering, and Noise Removal.} For long-form audio—particularly in unlabeled or newly collected data—we applied silence- and energy-based segmentation to break recordings into utterances. We retained segments with durations between 1 and 20 seconds to avoid instability caused by very short or excessively long samples. To further enhance quality, we removed segments with excessive background noise using energy-based filters. Additionally, we applied voice activity detection (VAD) and speaker diarization using pretrained pipelines from the \texttt{pyannote-audio} library, including the voice Activity detection~\cite{Bredin2020, Bredin2021} and speaker diarization~\cite{Plaquet23, Bredin23} models.

\paragraph{Metadata Consolidation.} All processed datasets were reformatted into a unified JSON-based schema compatible with the Hugging Face \texttt{datasets} library~\cite{lhoest2021datasets} and \texttt{fairseq} framework. Each entry includes metadata fields such as audio path, transcription (if available), language ID, dataset origin, and usage split.

\section{Baseline Models}\label{appdx_sec:baseline_models}

\begin{table*}[!ht]
\centering
\resizebox{0.99\textwidth}{!}{%
\begin{tabular}{lllccccccH}
\toprule
\textbf{Type} & \textbf{Language} & \textbf{ISO-3} & \textbf{Whisper-v3} & \textbf{M4T-v2} & \textbf{MMS-1B-All} & \textbf{AfriHubert} & \textbf{mHubert} & \textbf{XLS-R} & \textbf{Ours} \\
\midrule

\multirow{4}{*}{\textbf{African}} 
& Afrikaans & afr & \sttag & \sttag & \sttag \ldtag & \pttag & \pttag & \pttag & \sttag \tstag \ldtag \\
& Akuapim-twi & Akuapim-twi & -- & -- & -- & \pttag & -- & -- & \sttag \tstag \\
& Amharic & amh & \sttag & \sttag & \sttag \tstag \ldtag & \pttag & \pttag & \pttag & \sttag \ldtag \\
& Asante-twi & Asante-twi & -- & -- & -- & \pttag & -- & -- & \sttag \tstag \\
& Basaa & bas & -- & -- & \sttag \ldtag & -- & -- & -- & \sttag \\
& Bemba & bem & -- & -- & \sttag \tstag \ldtag & \pttag & -- & -- & \sttag \ldtag \\
& Central Atlas Tamazight & tzm & -- & -- & \ldtag & -- & -- & -- & \ldtag \\
& Dyula & dyu & -- & -- & \sttag \tstag \ldtag & -- & -- & -- & \sttag \\
& Edo & bin & -- & -- & \ldtag & -- & -- & -- & \ldtag \\
& Ewe & ewe & -- & -- & \sttag \tstag \ldtag & \pttag & -- & -- & \tstag \\
& Fanti & fat & -- & -- & -- & -- & -- & -- & \sttag \\
& Fon & fon & -- & -- & \sttag \tstag \ldtag & -- & -- & -- & \sttag \\
& Ga & gaa & -- & -- & \ldtag & -- & -- & -- & \sttag \\
& Ganda & lug & -- & \sttag & \sttag \tstag \ldtag & \pttag & \pttag & \pttag & \sttag \\
& Guerze/Kpelle & kpe & -- & -- & -- & -- & -- & -- & \ldtag \\
& Hausa & hau & \sttag & -- & \sttag \tstag \ldtag & \pttag & \pttag & \pttag & \sttag \tstag \ldtag \\
& Ibibio & ibb & -- & -- & \ldtag & -- & -- & -- & \ldtag \\
& Igbo & ibo & -- & \sttag & \sttag \ldtag & \pttag & \pttag & -- & \sttag \ldtag \\
& Kabyle & kab & -- & -- & \sttag \tstag \ldtag & -- & \pttag & \pttag & \sttag \\
& Kalenjin & kln & -- & -- & -- & -- & -- & -- & \sttag \\
& Kinyarwanda & kin & -- & -- & \sttag \tstag \ldtag & \pttag & \pttag & \pttag & \sttag \tstag \ldtag \\
& Kisi & kss & -- & -- & \sttag \tstag \ldtag & \pttag & -- & -- & \ldtag \\
& Konyanka Maninka & mku & -- & -- & \ldtag & \pttag & -- & -- & \ldtag \\
& Lingala & lin & \sttag & -- & \sttag \ldtag & \pttag & \pttag & -- & \tstag \ldtag \\
& Lozi & loz & -- & -- & \ldtag & \pttag & -- & -- & \sttag \ldtag \\
& Lunda & lun & -- & -- & \ldtag & \pttag & -- & -- & \ldtag \\
& Luo (Kenya and Tanzania) & luo & -- & \sttag & \sttag \ldtag & -- & -- & -- & \sttag \\
& Malagasy & mlg & \sttag & -- & \sttag \tstag \ldtag & \pttag & -- & \pttag & \ldtag \\
& Mandinka & mnk & -- & -- & \sttag \tstag \ldtag & \pttag & -- & -- & \ldtag \\
& Nigerian Pidgin & pcm & -- & -- & \sttag \tstag \ldtag & -- & -- & -- & \ldtag \\
& North Ndebele & nde & -- & -- & \ldtag & -- & -- & -- & \ldtag \\
& Northern Sotho (Sepedi) & nso & -- & -- & \sttag \ldtag & \pttag & -- & -- & \sttag \\
& Nyanja & nya & -- & \sttag & \sttag \tstag \ldtag & \pttag & -- & -- & \sttag \ldtag \\
& Oromo & orm & -- & -- & \sttag \tstag \ldtag & -- & -- & -- & \ldtag \\
& Pulaar & fuc & -- & -- & -- & -- & -- & -- & \sttag \\
& Pular & fuf & -- & -- & -- & \pttag & -- & -- & \sttag \ldtag \\
& Serer & srr & -- & -- & \ldtag & \pttag & -- & -- & \sttag \\
& Shona & sna & \sttag & \sttag & \sttag \tstag \ldtag & \pttag & \pttag & \pttag & \ldtag \\
& Somali & som & \sttag & \sttag & \sttag \tstag \ldtag & \pttag & \pttag & \pttag & \ldtag \\
& South Ndebele & nbl & -- & -- & \ldtag & \pttag & -- & -- & \sttag \\
& Southern Sotho & sot & -- & -- & \ldtag & \pttag & \pttag & -- & \sttag \tstag \\
& Standard Moroccan Tamazight & zgh & -- & -- & -- & -- & -- & -- & \sttag \\
& Susu & sus & -- & -- & \sttag \tstag \ldtag & \pttag & -- & -- & \sttag \\
& Susuami & ssu & -- & -- & -- & -- & -- & -- & \ldtag \\
& Swahili & swa, swh & \sttag & -- & \sttag \tstag \ldtag & \pttag & \pttag & \pttag & \sttag \ldtag \\
& Swati & ssw & -- & -- & \ldtag & \pttag & -- & -- & \sttag \\
& Taita & dav & -- & -- & -- & -- & -- & -- & \sttag \\
& Tem & kdh & -- & -- & \sttag \tstag \ldtag & -- & -- & -- & \ldtag \\
& Tigre & tig & -- & -- & \ldtag & -- & \pttag & -- & \sttag \\
& Tigrinya & tir & -- & -- & \sttag \tstag \ldtag & -- & -- & -- & \sttag \ldtag \\
& Tiv & tiv & -- & -- & \ldtag & -- & -- & -- & \ldtag \\
& Tonga (Zambia) & toi & -- & -- & \ldtag & \pttag & -- & -- & \sttag \ldtag \\
& Tsonga & tso & -- & -- & \sttag \tstag \ldtag & \pttag & -- & -- & \sttag \\
& Tswana & tsn & -- & -- & \ldtag & \pttag & \pttag & -- & \sttag \tstag \\
& Twi & twi & -- & -- & -- & -- & -- & -- & \sttag \\
& Venda & ven & -- & -- & \ldtag & \pttag & -- & -- & \sttag \\
& Western Maninkakan & mlq & -- & -- & \ldtag & -- & -- & -- & \sttag \\
& Wolof & wol & -- & -- & \sttag \ldtag & \pttag & -- & -- & \sttag \ldtag \\
& Xhosa & xho & -- & \sttag & \sttag \ldtag & \pttag & \pttag & -- & \sttag \tstag \\
& Yoruba & yor & -- & \sttag & \sttag \tstag \ldtag & \pttag & \pttag & \pttag & \sttag \tstag \ldtag \\
& Zulu & zul & \sttag & \sttag & \sttag \ldtag & \pttag & -- & \pttag & \sttag \\
\midrule
\multirow{5}{*}{\textbf{Code-switched}} 
& English - Southern Sotho & eng-sot & -- & -- & -- & -- & -- & -- & \sttag \\
& English - Tswana & eng-tsn & -- & -- & -- & -- & -- & -- & \sttag \\
& English - Xhosa & eng-xho & -- & -- & -- & -- & -- & -- & \sttag \\
& English - Zulu & eng-zul & -- & -- & -- & -- & -- & -- & \sttag \\
& Northern Sotho - English & nso-eng & -- & -- & -- & -- & -- & -- & \sttag \\
\midrule
\textbf{Non-African} & English - Accented & eng & \sttag & \sttag & \sttag \tstag \ldtag & \pttag & \pttag & \pttag & \sttag \ldtag \\

\bottomrule
\end{tabular}%
}
\caption{Overview of African language coverage across models for pretraining and downstream speech and language tasks. \textbf{Abbreviations}: \sttag~Speech-to-Text (ASR), \tstag~Text-to-Speech, \ldtag~Spoken Language Identification, and \pttag~Pretraining.}

\label{appdx_tab:model_support_types}
\end{table*}

\paragraph{Whisper-v3}~\cite{whisper}: Developed by OpenAI, Whisper-v3 is a large-scale encoder-decoder model trained on 680k hours of multilingual and multitask supervised data. We evaluate two variants: \texttt{whisper-large-v3}, which offers high accuracy for multilingual ASR tasks, and \texttt{whisper-large-v3-turbo}, which provides faster inference with a slight trade-off in accuracy.

\paragraph{SeamlessM4T-v2 Large}~\cite{seamlessm4t2023}: A unified model by Meta AI supporting speech-to-text, speech-to-speech, and text-to-text translation across over 100 languages. It is particularly designed for low-latency and zero-shot multilingual translation.

\paragraph{MMS-1b-All}~\cite{pratap2023mms}: Part of Meta’s Massively Multilingual Speech (MMS) project, this model is trained on over 1,100 languages with 1B parameters. It supports  ASR and SLID tasks and represents the largest multilingual speech pretraining effort to date. ~\citeauthor{pratap2023mms} also trained the VITs architecture for TTS on a number of languages, including 3 of the African languages covered in our work.

\paragraph{AfriHUBERT}~\cite{alabi2024afrihubert}: A HuBERT-based self-supervised model trained exclusively on African speech data. It focuses on improving representation learning for low-resource African languages and is optimized for ASR and feature extraction tasks.

\paragraph{Wav2Vec2-XLS-R}~\cite{babu2021xls}: A family of cross-lingual speech representation models developed by Facebook AI, trained using the wav2vec2 framework on a multilingual dataset spanning 128 languages. We evaluate two variants: \texttt{facebook/wav2vec2-xls-r-300m} and \texttt{facebook/wav2vec2-xls-r-1b}, which differ in parameter count and pretraining scale. These models are widely used for fine-tuning on low-resource ASR tasks due to their strong generalization across languages.

Table~\ref{appdx_tab:model_support_types} presents a detailed overview of African language support across models for pretraining and various downstream tasks in speech and language processing.

\section{Experimental Setup}\label{appdx_sec:hyparam}
For the \ourmodel series of models, we select the best checkpoint for each model based on development set performance at the end of each epoch. These selected checkpoints are then used for final evaluation, during which we compute task-specific metrics and report the results accordingly.

\paragraph{Hyperparameters.}
All ASR and SLID models are fine-tuned using the Adam optimizer with a cosine learning rate of $5 \times 10^{-5}$ over 30 epochs. We use the HuggingFace Transformers~\cite{wolf-etal-2020-transformers} for training and evaluation.\footnote{\href{https://github.com/huggingface/transformers}{https://github.com/huggingface/transformers}}  For TTS models, we adopt the finetuning procedure outlined in the Vits repository and follow the default hyperparameter configuration provided in the repository\footnote{\href{https://github.com/ylacombe/finetune-hf-vits}{https://github.com/ylacombe/finetune-hf-vits}}.
\paragraph{Evaluation Metrics.}
For ASR, we evaluate using Word Error Rate (WER) and Character Error Rate (CER). For TTS, we assess synthesized speech intelligibility with the best available ASR model for each language, reporting both WER and CER as objective measures following \citet{toyin-etal-2023-artst}. For SLID, we use macro-F\textsubscript{1} to address class imbalance and ensure balanced performance assessment across languages.

\section{Evaluation Results}\label{appdx_sec:results}

Table~\ref{appdx_table:asr_results_full} shows the detailed results of all models across all AST test sets.  
Table~\ref{appdx_tab:slid-results} presents the results for the spoken language identification task.

\begin{table*}[!ht]
\centering
\resizebox{0.99\textwidth}{!}{%
\begin{tabular}{llrrrrrrrrrrH}
\toprule
\multicolumn{1}{c}{} & \multicolumn{1}{c}{} & &&& & \multicolumn{5}{c}{\textbf{\ourmodel Series (\textcolor{red}{Ours})}} & \multicolumn{1}{c}{} \\
 \cmidrule{8-12}
\multicolumn{1}{c}{\multirow{-2}{*}{\textbf{Language}}} & 
\multicolumn{1}{c}{\multirow{-2}{*}{\textbf{Test Set}}} & 
\multirow{-2}{*}{\textbf{MMS}} & \multirow{-2}{*}{\textbf{Seamless}} & \multirow{-2}{*}{\textbf{Whisper}} & \multirow{-2}{*}{\textbf{WhisperT}} & & 
\textbf{\ourmodel-H} & \textbf{\ourmodel-M} & \textbf{\ourmodel-S} & \textbf{\ourmodel-X} & \textbf{\ourmodel-W} & 
\multicolumn{1}{H}{\multirow{-2}{*}{\textbf{\ourmodel}}} \\
\midrule

Akuapim-twi (aka) &FS&85.82/40.14&\linecolor{red!100}{219.67/190.49}&\linecolor{red!100}{1181.0/1131.23}&\linecolor{red!100}{499.51/547.24}&&26.83/10.13&17.6/8.13&\colorbox{green!20}{13.29/8.45}&23.74/10.35&29.1/19.1 & 36.64/22.64\\ \midrule
Asante-twi (aka)&FS&83.6/32.35&\linecolor{red!100}{230.88/196.71}&\linecolor{red!100}{665.34/574.27}&\linecolor{red!100}{245.5/222.37}&&26.78/7.36&13.87/5.38&\colorbox{green!20}{7.06/2.62}&19.93/7.06&15.63/7.98& 27.09/14.44\\ \midrule
Afrikaans (afr)&Lwazi&92.06/37.59&37.91/16.47&66.05/34.32&73.17/39.05&&62.81/17.9&36.29/9.86&\colorbox{green!20}{15.62/4.99}&102.96/53.45&29.22/11.0&38.09/12.46\\
Afrikaans (afr)&NCHTL&118.72/31.86&27.96/4.63&77.61/24.22&67.61/15.2&&53.57/8.16&25.55/3.4&\colorbox{green!20}{12.39/2.01}&109.93/36.25&20.82/3.81&51.35/8.26\\
Afrikaans (afr)&CV-19&26.29/6.7&19.52/9.18&35.85/9.9&46.38/17.11&&64.15/19.97&35.36/13.19&\colorbox{green!20}{16.97/7.47}&93.32/46.55&27.87/11.27&98.29/44.35\\ \midrule
Amharic (amh)&CV-19&51.93/21.81&87.58/22.25&432.1/294.11&245.47/236.28&&86.93/42.59&58.26/25.39&\colorbox{green!20}{42.14/16.94}&105.96/119.54&106.34/65.09& 78.68/54.68\\ \midrule
 Basaa (bas)&CV-19&34.4/9.6&\linecolor{red!100}{147.17/109.79}&\linecolor{red!100}{554.16/475.04}&\linecolor{red!100}{169.5/123.55}&&61.08/20.41&\colorbox{green!20}{36.51/10.27}&65.17/24.97&84.09/30.86&76.39/31.3&92.81/45.13\\ \midrule
 Bemba (bem)&BS&47.73/7.95&\linecolor{red!100}{187.48/106.1}&\linecolor{red!100}{921.43/515.91}&\linecolor{red!100}{136.53/70.37}&&51.9/9.28&44.06/7.1&\colorbox{green!20}{38.99/7.59}&83.32/20.12&50.84/10.51&70.11/23.23\\ \midrule
 Taita (dav)&CV-19&\linecolor{red!100}{82.47/25.19}&\linecolor{red!100}{170.25/104.12}&\linecolor{red!100}{662.71/401.46}&\linecolor{red!100}{151.56/86.05}&&67.34/20.59&58.49/16.99&\colorbox{green!20}{44.79/15.29}&82.66/27.59&105.83/60.98&84.25/33.49\\ \midrule
 Dyula (dyu)&CV-19&65.61/16.14&\linecolor{red!100}{152.07/104.41}&\linecolor{red!100}{424.53/344.84}&\linecolor{red!100}{107.85/43.71}&&77.98/23.26&\colorbox{green!20}{67.99/21.53}&78.07/23.11&85.58/26.57&87.02/26.42&79.38/43.69\\ \midrule
 
 Fanti (fat)&FS&\linecolor{red!100}{115.34/61.98}&\linecolor{red!100}{244.53/209.09}&\linecolor{red!100}{1188.25/1082.92}&\linecolor{red!100}{497.67/581.04}&&23.38/7.27&19.97/6.99&\colorbox{green!20}{8.58/4.96}&27.89/9.94&23.06/15.66&26.43/12.72\\ \midrule
 Fon (fon)&Alffa&87.83/31.92&\linecolor{red!100}{132.67/115.18}&\linecolor{red!100}{488.58/467.05}&\linecolor{red!100}{159.01/134.6}&&\colorbox{green!20}{33.29/9.49}&44.51/12.52&43.75/14.77&53.81/17.4&45.54/16.72&43.31/21.28\\ \midrule
 Pulaar (fuc)&Kallaama&\linecolor{red!100}{103.25/68.15}&\linecolor{red!100}{200.49/144.98}&\linecolor{red!100}{904.99/743.08}&\linecolor{red!100}{321.82/280.08}&&91.74/56.75&87.29/54.54&\colorbox{green!20}{69.39/43.09}&96.67/67.88&107.04/73.33&\\ \midrule

 Pular (fuf)&NL4-WA&\linecolor{red!100}{106.98/51.34}&\linecolor{red!100}{244.57/177.0}&\linecolor{red!100}{789.15/740.44}&\linecolor{red!100}{553.1/435.86}&&106.2/50.68&101.55/44.86&98.06/54.05&\colorbox{green!20}{96.9/53.01}&136.05/75.85&95.77/58.53\\ \midrule

 Ga (gaa)&FS&\linecolor{red!100}{139.31/55.26}&\linecolor{red!100}{322.68/230.61}&\linecolor{red!100}{1362.33/1043.95}&\linecolor{red!100}{482.59/412.01}&&41.56/11.51&20.35/7.35&\colorbox{green!20}{9.67/6.38}&32.97/10.37&22.21/11.99&24.89/13.15\\ \midrule

Hausa (hau)&CV-19&27.63/5.97&\linecolor{red!100}{135.46/91.76}&110.59/56.15&130.84/72.4&&55.68/15.58&\colorbox{green!20}{29.42/6.58}&64.19/23.1&92.06/34.61&90.54/47.26&95.02/46.11\\ \midrule
 Igbo (ibo)&CV-19&70.82/18.08&61.66/18.02&\linecolor{red!100}{111.67/62.92}&\linecolor{red!100}{321.93/175.32}&&89.02/35.57&83.35/26.58&\colorbox{green!20}{77.73/34.93}&98.33/42.57&95.27/54.04&95.00/44.78\\ \midrule
Kabyle (kab)&CV-19&49.49/14.33&\linecolor{red!100}{149.33/101.52}&\linecolor{red!100}{508.87/412.14}&\linecolor{red!100}{153.96/123.25}&&79.21/25.64&62.81/16.99&\colorbox{green!20}{58.78/20.46}&93.46/44.89&67.02/29.09&\\ \midrule
 Kinyarwanda (kin)&CV-19&34.22/9.42&\linecolor{red!100}{167.9/93.77}&\linecolor{red!100}{820.55/473.72}&\linecolor{red!100}{245.92/141.28}&&55.33/15.82&\colorbox{green!20}{38.59/10.23}&54.22/18.14&91.2/33.29&72.8/24.77&\\
Kalenjin (kln)&CV-19&\linecolor{red!100}{99.97/34.64}&\linecolor{red!100}{178.09/95.33}&\linecolor{red!100}{773.64/453.35}&\linecolor{red!100}{178.81/107.16}&&80.44/21.49&73.43/18.86&\colorbox{green!20}{70.37/18.19}&85.26/25.65&75.93/21.24&\\ \midrule
 Lozi (loz)&Z.Voice&\linecolor{red!100}{87.37/32.2}&\linecolor{red!100}{124.72/95.9}&\linecolor{red!100}{657.46/508.83}&\linecolor{red!100}{109.26/55.17}&&61.27/23.84&63.58/23.66&\colorbox{green!20}{57.34/22.92}&87.28/32.05&64.39/24.12&\\ \midrule
 Ganda (lug)&CV-19&26.21/5.35&17.69/4.27&\linecolor{red!100}{866.81/468.58}&\linecolor{red!100}{168.18/77.31}&&64.15/13.64&35.24/6.37&\colorbox{green!20}{23.11/5.65}&88.66/25.19&55.92/13.83&\\ \midrule
Luo (luo)&CV-19&111.02/76.27&111.43/53.86&\linecolor{red!100}{478.84/332.08}&\linecolor{red!100}{115.16/53.15}&&56.86/13.6&42.4/9.05&\colorbox{green!20}{38.79/10.29}&67.28/16.55&52.18/13.27&\\ \midrule
 W. Maninkakan (mlq)&NL4-WA&\linecolor{red!100}{113.02/59.21}&\linecolor{red!100}{228.93/171.01}&\linecolor{red!100}{1232.92/1237.86}&\linecolor{red!100}{306.11/217.65}&&110.62/48.96&98.97/40.52&115.82/51.12&\colorbox{green!20}{96.65/47.21}&176.76/113.79&\\ \midrule
 S. Ndebele (nbl)&Lwazi&\linecolor{red!100}{74.29/31.76}&\linecolor{red!100}{139.42/91.29}&\linecolor{red!100}{349.57/199.2}&\linecolor{red!100}{100.43/56.06}&&62.13/18.33&38.44/10.89&\colorbox{green!20}{19.02/7.4}&103.06/52.98&29.58/11.33&\\
 S. Ndebele (nbl)&NCHTL&\linecolor{red!100}{58.61/10.53}&\linecolor{red!100}{238.25/104.08}&\linecolor{red!100}{1368.24/566.76}&\linecolor{red!100}{198.86/86.4}&&31.95/5.57&33.13/5.45&\colorbox{green!20}{25.51/4.99}&66.75/11.16&36.32/6.14&\\ \midrule
 Northern Sotho (nso)&Lwazi&84.64/32.4&\linecolor{red!100}{147.45/94.85}&\linecolor{red!100}{251.7/175.66}&\linecolor{red!100}{105.29/71.63}&&67.06/19.27&43.43/11.37&\colorbox{green!20}{21.27/7.84}&104.06/54.47&33.07/10.22&\\
 N. Sotho (nso)&NCHTL&42.69/11.39&\linecolor{red!100}{154.46/120.61}&\linecolor{red!100}{611.95/512.8}&\linecolor{red!100}{158.31/140.4}&&20.72/5.21&21.49/5.09&\colorbox{green!20}{16.39/4.42}&47.05/13.71&22.45/6.44&\\ \midrule
 Nyanja (nya)&Z.Voice&99.85/82.25&25.34/7.0&\linecolor{red!100}{744.72/392.55}&\linecolor{red!100}{92.22/23.12}&&50.61/10.99&46.8/9.78&\colorbox{green!20}{22.38/5.99}&76.22/18.17&41.61/8.94&\\ \midrule
 S. Sotho (sot)&Lwazi&\linecolor{red!100}{70.04/29.11}&\linecolor{red!100}{132.03/86.73}&\linecolor{red!100}{248.55/193.15}&\linecolor{red!100}{110.71/52.4}&&61.59/17.94&38.2/10.41&\colorbox{green!20}{18.63/7.24}&102.48/54.0&31.81/11.55&\\
 S. Sotho (sot)&NCHTL&\linecolor{red!100}{79.97/27.48}&\linecolor{red!100}{154.26/111.44}&\linecolor{red!100}{743.88/591.26}&\linecolor{red!100}{145.42/113.09}&&23.94/6.31&26.84/6.87&\colorbox{green!20}{18.15/5.58}&44.74/12.54&24.47/7.3&\\ \midrule

 Serer (srr)&Kallaama&\linecolor{red!100}{105.41/69.85}&\linecolor{red!100}{255.33/233.38}&\linecolor{red!100}{1046.88/977.99}&\linecolor{red!100}{479.84/571.07}&&95.21/55.41&94.26/56.94&\colorbox{green!20}{88.39/62.34}&96.22/68.31&125.44/113.36&\\ \midrule

 Swati (ssw)&Lwazi&\linecolor{red!100}{73.08/29.37}&\linecolor{red!100}{139.27/88.48}&\linecolor{red!100}{338.16/309.79}&\linecolor{red!100}{113.7/61.78}&&64.93/18.42&39.59/10.4&\colorbox{green!20}{17.94/6.57}&101.49/54.47&30.79/11.01&\\
 Swati (ssw)&NCHTL&\linecolor{red!100}{65.0/10.76}&\linecolor{red!100}{247.32/106.42}&\linecolor{red!100}{1345.67/539.36}&\linecolor{red!100}{221.86/77.57}&&22.88/3.15&29.39/4.14&\colorbox{green!20}{20.6/3.28}&62.45/9.55&34.35/5.07&\\ \midrule
 Susu (sus)&NL4-WA&150.79/123.0&\linecolor{red!100}{264.17/177.19}&\linecolor{red!100}{665.0/471.49}&\linecolor{red!100}{491.35/496.48}&&120.4/48.53&\colorbox{green!20}{107.5/36.83}&126.55/51.74&108.81/44.54&215.16/121.32&\\ \midrule
 Swahili (swa)&CV-19&25.65/7.46&15.86/6.16&81.89/38.84&95.0/42.51&&42.46/11.6&24.77/7.23&\colorbox{green!20}{16.52/6.15}&68.07/19.14&34.7/11.87&\\
 Swahili (swh)&Alffa&40.8/12.37&25.0/10.47&63.9/23.29&63.55/24.44&&43.87/11.36&29.29/7.87&\colorbox{green!20}{16.61/5.64}&71.51/22.25&25.84/8.22&\\ \midrule
Tigre (tig)&CV-19&\linecolor{red!100}{115.07/120.83}&\linecolor{red!100}{213.66/207.69}&\linecolor{red!100}{690.76/652.88}&\linecolor{red!100}{143.04/179.65}&&71.94/30.13&59.25/21.02&\colorbox{green!20}{57.74/26.16}&102.46/90.21&87.39/67.43&\\ \midrule
Tigrinya (tir)&CV-19&117.74/111.01&\linecolor{red!100}{189.88/180.49}&\linecolor{red!100}{165.36/148.48}&\linecolor{red!100}{135.48/158.64}&&90.24/47.95&92.5/65.98&\colorbox{green!20}{75.24/50.84}&100.71/91.24&122.5/115.54&\\ \midrule
 Tonga (Zambia) (toi)&Z.Voice&\linecolor{red!100}{71.71/14.91}&\linecolor{red!100}{188.65/92.13}&\linecolor{red!100}{1175.58/550.42}&\linecolor{red!100}{127.14/41.37}&&63.02/10.74&\colorbox{green!20}{42.25/6.82}&51.31/8.01&85.57/22.14&57.49/10.66&\\ \midrule
 Tswana (tsn)&Lwazi&\linecolor{red!100}{72.4/31.33}&\linecolor{red!100}{140.76/93.65}&\linecolor{red!100}{231.9/159.83}&\linecolor{red!100}{119.46/84.15}&&62.14/17.09&37.45/10.51&\colorbox{green!20}{18.2/6.64}&102.84/53.11&28.44/10.2&\\
 Tswana (tsn)&NCHTL&\linecolor{red!100}{63.26/18.88}&\linecolor{red!100}{165.25/109.5}&\linecolor{red!100}{795.5/551.5}&\linecolor{red!100}{161.95/135.85}&&18.9/4.26&22.38/4.88&\colorbox{green!20}{12.95/3.46}&44.1/10.84&18.86/4.87&\\ \midrule
 Tsonga (tso)&Lwazi&80.41/33.76&\linecolor{red!100}{142.02/91.96}&\linecolor{red!100}{264.92/172.82}&\linecolor{red!100}{91.92/55.2}&&62.69/18.33&38.48/9.98&\colorbox{green!20}{17.0/6.05}&102.67/53.21&34.21/20.76&\\
 Tsonga (tso)&NCHTL&61.74/10.46&\linecolor{red!100}{163.28/107.2}&\linecolor{red!100}{1105.49/748.39}&\linecolor{red!100}{148.77/102.28}&&22.5/4.0&25.87/4.41&\colorbox{green!20}{17.77/3.65}&55.94/11.75&27.45/6.12&\\ \midrule
 Twi (twi)&CV-19&\linecolor{red!100}{94.32/42.27}&\linecolor{red!100}{128.78/96.52}&\linecolor{red!100}{599.6/403.21}&\linecolor{red!100}{105.51/52.47}&&74.97/26.69&81.06/30.01&\colorbox{green!20}{62.68/15.21}&84.81/31.96&91.34/28.7&\\ \midrule
 Venda (ven)&Lwazi&\linecolor{red!100}{71.16/29.89}&\linecolor{red!100}{140.9/95.47}&\linecolor{red!100}{265.76/220.81}&\linecolor{red!100}{129.56/155.71}&&62.66/19.31&38.71/11.41&\colorbox{green!20}{19.13/6.96}&102.19/52.9&30.95/11.58&\\
 Venda (ven)&NCHTL&\linecolor{red!100}{85.98/27.41}&\linecolor{red!100}{159.21/112.41}&\linecolor{red!100}{653.82/456.88}&\linecolor{red!100}{122.93/79.61}&&28.28/6.12&33.11/6.99&\colorbox{green!20}{27.37/6.89}&68.34/20.29&32.21/7.87&\\ \midrule
 Wolof (wol)&Alffa&43.57/10.44&\linecolor{red!100}{128.76/92.3}&\linecolor{red!100}{446.07/348.0}&\linecolor{red!100}{202.6/143.81}&&59.7/15.41&34.75/8.03&40.65/13.28&89.3/30.42&\colorbox{green!20}{34.42/9.82}&\\

 Wolof (wol)&Kallaama&101.14/81.39&\linecolor{red!100}{1050.1/1020.36}&\linecolor{red!100}{1050.1/1020.36}&\linecolor{red!100}{374.1/388.4}&&105.0/75.1&\colorbox{green!20}{100.44/77.09}&100.20/75.38&102.95/82.07&143.16/131.49&\\ \midrule

 Xhosa (xho)&Lwazi&73.89/33.13&140.48/90.83&\linecolor{red!100}{286.14/227.12}&\linecolor{red!100}{148.78/78.46}&&67.97/19.69&43.26/11.49& 22.1/7.83 &101.99/53.45&46.67/38.59&\\
 Xhosa (xho)&NCHTL&35.24/5.76&246.81/117.89&\linecolor{red!100}{1405.1/615.03}&\linecolor{red!100}{217.5/87.08}&&34.43/5.58&32.33/5.09&\colorbox{green!20}{28.66/5.28}&68.16/11.29&40.82/7.08&\\ \midrule
 Yoruba (yor)&Y.Voice&50.12/18.29&23.47/11.8&\linecolor{red!100}{639.75/503.98}&\linecolor{red!100}{105.84/70.0}&&40.59/13.49&41.21/12.91&\colorbox{green!20}{20.12/11.72}&98.51/55.74&52.01/25.45&\\ \midrule
S. M. Tamazight (zgh)&CV-19&\linecolor{red!100}{107.34/98.24}&\linecolor{red!100}{150.25/123.89}&\linecolor{red!100}{371.93/326.05}&\linecolor{red!100}{129.36/125.21}&&102.04/86.11&\colorbox{green!20}{90.85/72.04}&111.43/98.69&101.33/91.43&108.25/94.8&\\ \midrule
 Zulu (zul)&Lwazi&70.12/32.66&107.96/84.77&164.54/106.64&78.11/43.35&&62.92/17.57&38.58/10.88&108.53/103.61&101.93/52.87&\colorbox{green!20}{27.63/10.87}&\\
 Zulu (zul)&NCHTL&31.31/5.12&74.28/20.56&648.45/244.13&379.87/134.73&&30.55/4.69&26.36/3.96&\colorbox{green!20}{23.87/4.47}&60.96/8.79&33.92/5.71&\\ \midrule

\multicolumn{2}{c}{\textbf{Overall Average}} &75.9/35.26&146.69/98.92&611.91/437.98&196.7/149.79&&59.9/21.46&48.11/17.41&\colorbox{green!20}{41.65/18.3}&82.64/39.31&60.56/31.16&\\ \midrule

\end{tabular}%
}
\caption{Comparison of ASR performance across various African languages using baseline models and our's \ourmodel models in both zero-shot and fine-tuned settings. The evaluation metrics are reported as \texttt{WER/CER}. \linecolor{red}{Red Underline} indicates that the model does not support the corresponding language. \colorbox{green!20}{Green} indicates the best-performing model for each language/test set. \textbf{Abbreviations}: \texttt{FS} – Financial Speech, \texttt{BS} – Bemba Speech, \texttt{CV-19} – Common Voice 2019, \texttt{NL4-WA} – Nicolingua-0004-West Africa, \texttt{Z.Voice} – Zambezi Voice, \texttt{Y.Voice} – Yoruba Voice, \texttt{S.M.} – Standard Moroccan, \texttt{N.} – Northern, \texttt{S.} – South/Southern, \texttt{W.} – Westren.}
\label{appdx_table:asr_results_full} 
\end{table*}

\begin{table}[!ht]
\centering
\resizebox{0.99\columnwidth}{!}{%
\begin{tabular}{llrr}
\toprule
\textbf{Language}             & \textbf{Test Set} & \textbf{MMS-LID-1024} & \textbf{\ourmodel-SLID} \\
\midrule
Edo (bin)                     & OlogoAfrica      & 6.25                  & \colorbox{green!20}{80.12}               \\ \midrule
Afrikaans (afr)               & UDHR             & \colorbox{green!20}{88.89}                 & \colorbox{green!20}{88.89}               \\\midrule
Amharic (amh)                 & UDHR             & \colorbox{green!20}{100.00}                   & \colorbox{green!20}{100.00}                 \\
Amharic (amh)                 & VoxLingua        & \colorbox{green!20}{98.50}                 & 89.39               \\\midrule
Bemba (bem)                   & ZambeziVoice     & 26.67                 & \colorbox{green!20}{53.15 }              \\\midrule
Hausa (hau)                   & OlogoAfrica      & \colorbox{green!20}{100.00 }                  & \colorbox{green!20}{100.00 }                \\
Hausa (hau)                   & UDHR             & \colorbox{green!20}{75.00 }                   & \colorbox{green!20}{75.00 }                 \\
Hausa (hau)                   & VoxLingua        & \colorbox{green!20}{97.99}                 & 94.18               \\\midrule
Ibibio (ibb)                  & OlogoAfrica      & 14.29                 & \colorbox{green!20}{25.98}               \\\midrule
Igbo (ibo)                    & OlogoAfrica      & 65.79                 & \colorbox{green!20}{75.26}               \\\midrule
Tem (kdh)                     & UDHR             & 45.45                 & \colorbox{green!20}{55.23}               \\\midrule
Kinyarwanda (kin)             & VOA              & 20.50                  & \colorbox{green!20}{21.92}               \\\midrule
Lingala (lin)                 & VoxLingua        & \colorbox{green!20}{96.29}                 & 15.86               \\\midrule
Lozi (loz)                    & ZambeziVoice     & 1.36                  & \colorbox{green!20}{5.30}                 \\\midrule
Lunda (lun)                   & ZambeziVoice     & 23.60                  & \colorbox{green!20}{30.12}               \\\midrule
Malagasy (mlg)                & VoxLingua        & \colorbox{green!20}{98.55}                 & 70.12               \\\midrule
Nyanja (nya)                  & ZambeziVoice     & \colorbox{green!20}{22.64}                 & 15.59               \\\midrule
Nigerian Pidgin (pcm)         & OlogoAfrica      & 73.68                 & \colorbox{green!20}{74.32}               \\\midrule
Shona (sna)                   & OlogoAfrica      & 90.91                 & \colorbox{green!20}{92.34}               \\
Shona (sna)                   & VoxLingua        & 86.61                 & \colorbox{green!20}{88.23}               \\\midrule
Somali (som)                  & VoxLingua        & \colorbox{green!20}{97.96}                 & 95.54               \\\midrule
Swahili (swa, swh)            & OlogoAfrica      & \colorbox{green!20}{99.03}                 & 94.14               \\
Swahili (swa, swh)            & UDHR             & \colorbox{green!20}{99.60}                  & 94.29               \\
Swahili (swa, swh)            & VoxLingua        & \colorbox{green!20}{99.96}                 & 94.29               \\\midrule
Tiv (tiv)                     & OlogoAfrica      & 66.67                 & \colorbox{green!20}{69.93}               \\\midrule
Tonga (Zambia) (toi)          & ZambeziVoice     & 31.48                 & \colorbox{green!20}{56.47}               \\\midrule
Central Atlas Tamazight (tzm) & OlogoAfrica      & 27.27                 & \colorbox{green!20}{40.76}               \\\midrule
Wolof (wol)                   & UDHR             & \colorbox{green!20}{83.33}                 & \colorbox{green!20}{83.33}               \\\midrule
Yoruba (yor)                  & OlogoAfrica      & \colorbox{green!20}{100.00}                   & \colorbox{green!20}{100.00}                 \\
Yoruba (yor)                  & VoxLingua        & \colorbox{green!20}{96.27}                 & 95.87               \\ \bottomrule
\multicolumn{2}{l}{\textbf{Overall Average}}                      & 69.44                 & \colorbox{green!20}{70.82}              \\ \bottomrule
\end{tabular}%
}

\caption{Performance of the MMS-LID-1024 on \ourbenchmark and \ourmodel-SLID. \colorbox{green!20}{Green} indicates the best-performing model for each language/test set. The evaluation metrics are reported as \texttt{$F_1-macro$}.}
\label{appdx_tab:slid-results}
\end{table}

\end{document}